\RequirePackage{fix-cm}
\documentclass[twocolumn]{svjour3}          

\usepackage{graphics} 
\usepackage{graphicx} 
\usepackage{amsmath} 
\usepackage{amssymb}  
\usepackage{hyperref}
\usepackage{float}
\usepackage{cite} 
\usepackage[ruled, lined]{algorithm2e} 
\usepackage[shortcuts]{extdash}

\usepackage{color}

\journalname{Autonomous Robots}
\begin{document}
\title{Autonomously Learning to Visually Detect Where Manipulation Will Succeed}

\author{ Hai Nguyen, and Charles C. Kemp}

\institute{
              Healthcare Robotics Lab,\\
              Georgia Institute of Technology \\
              Atlanta, GA, USA\\
              \email{haidai@gmail.com}\\
              \email{charlie.kemp@bme.gatech.edu}
       }

\maketitle


\begin{abstract}
Visual features can help predict if a manipulation behavior will
succeed at a given location. For example, the success of a behavior
that flips light switches depends on the location of the
switch. Within this paper, we present methods that enable a mobile
manipulator to autonomously learn a function that takes an RGB image
and a registered 3D point cloud as input and returns a 3D location at
which a manipulation behavior is likely to succeed.  Given a pair of
manipulation behaviors that can change the state of the world between
two sets (e.g., light switch up and light switch down), classifiers
that detect when each behavior has been successful, and an initial
hint as to where one of the behaviors will be successful, the robot
autonomously trains a pair of support vector machine (SVM) classifiers
by trying out the behaviors at locations in the world and observing
the results. When an image feature vector associated with a 3D
location is provided as input to one of the SVMs, the SVM predicts if
the associated manipulation behavior will be successful at the 3D
location. To evaluate our approach, we performed experiments with a
PR2 robot from Willow Garage in a simulated home using behaviors that
flip a light switch, push a rocker-type light switch, and operate a
drawer. By using active learning, the robot efficiently learned SVMs
that enabled it to consistently succeed at these tasks. After
training, the robot also continued to learn in order to adapt in the
event of failure.
\end{abstract}
\section{Introduction}

\begin{figure} [t]
\centering
\includegraphics[height=7.0cm]{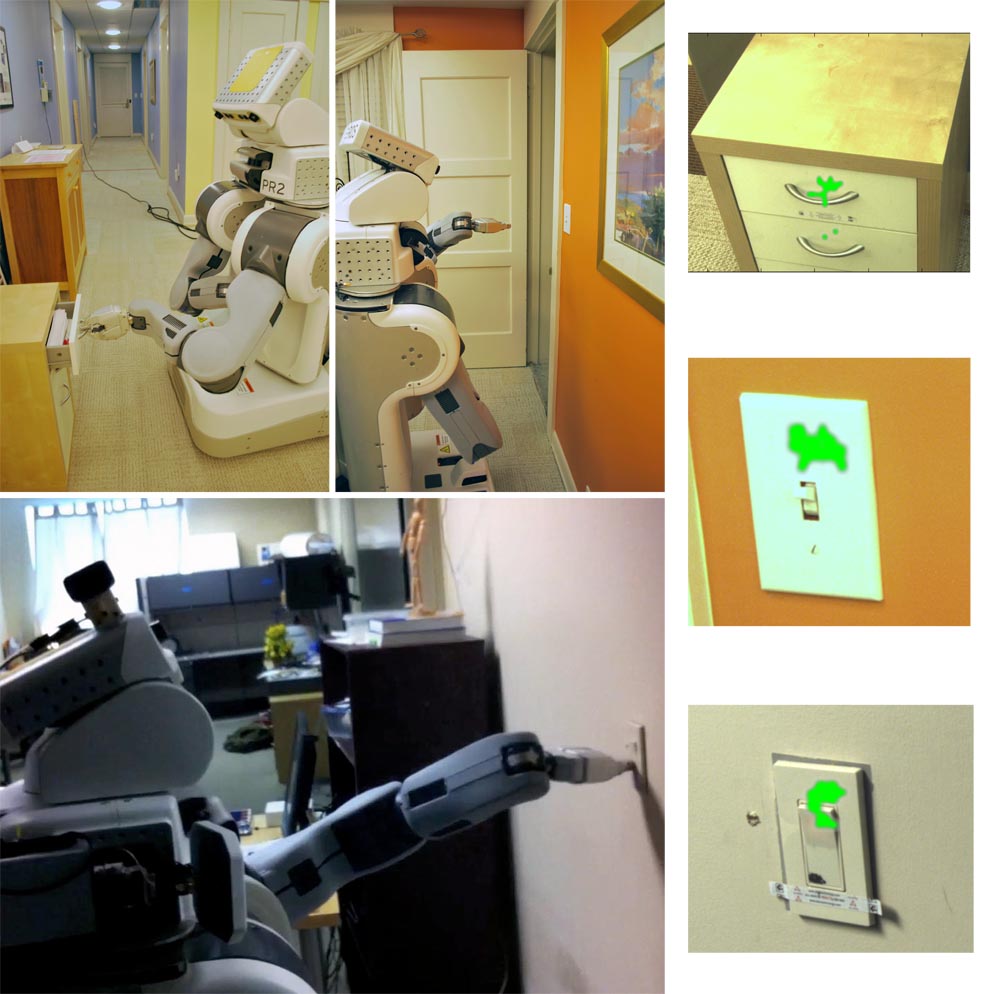}
\caption{\label{fig_teaser}
\textbf{Left:} Willow Garage PR2 operating a drawer, light switch and rocker
switch using learned detector that detects regions where manipulation
will succeed.  
\textbf{Right:} Results from learned detectors during execution.
}
\end{figure}

Informing robot manipulation with computer vision continues to be a
challenging problem in unstructured human environments, such as
homes. Two types of challenges are particularly notable. First, the
robot must handle wide variation in the appearance of task-relevant
components of the world that can affect its ability to perform tasks
successfully. Lighting can vary from home to home and from hour to
hour due to indoor lighting and windows.  In addition, important
components, such as drawer handles and the drawer faces that serve as
background, can be distinctive or even unique. The perspective from
which a mobile robot observes the component will also vary.

Second, the relationship between the appearance of task-relevant
components and the success or failure of a manipulation behavior is
complex. For example, the mechanics of a specific device may
require that the robot act at a distinct location, such as a finicky
drawer that needs to be pushed in the center to be closed, or a
convoluted handle that the robot's gripper can only grasp at
particular locations. The robot itself may also change over time and
thus alter the relationship between visual appearance and a
manipulation behavior, as parts of its body settle, deform, and wear.

One potential solution to these two problems is for robots to
autonomously learn how specific objects respond to manipulation
attempts using a behavior, and to continue to learn as they perform
tasks. By using self-generated data, robots can learn direct mappings
from visual features to the input parameters for behaviors, enabling
robust execution despite errors in calibration, variations in robot
pose, sensor noise, unexpected environmental interactions, and other
factors.  By continuing to learn over time, robots can also adapt to
changes in the environment, the objects, and their bodies.

In this work, we present a system that enables mobile manipulators to
autonomously gather data about the execution of behaviors to improve
their likelihood of success in future attempts. Our work advances
autonomous robot learning in three ways. First, our research addresses
challenges of learning in scenarios that integrate mobility and
manipulation. During our tests, the robot navigates to the device from
various places in the environment. Our approach uses a robot's
mobility as an integral part of autonomous learning, which enables the
robot to handle the significant task variation introduced by its
mobility. Second, we show that autonomously learning to visually
predict where a behavior will be successful can be tractable in
real-world scenarios. By using active learning, the robots in our
tests learned each visual function after fewer than 150 interactions
with each device, even though the robot started from scratch and only
used data it collected. The learned visual functions enabled the
robots to successfully operate the devices and also have intuitive
interpretations. Third, our methods autonomously learn to operate
devices that have an approximately binary state, such as a light
switch being up or down or a drawer being open or closed. This
presents a challenge, since the robot's actions change the state of
the world, which deters the robot from trying the same action
again. For example, it would be difficult to learn to open a drawer
if, once it is open, the robot is unable to close it.  Our system
addresses this difficulty by simultaneously training pairs of
behaviors and alternating between them as necessary. We also formalize
the ideal relationship between these pairs of behaviors and name them
complementary behaviors.

We evaluated our system using an implementation on a Willow Garage PR2
robot \cite{willow_website} at the Aware Home, which is a
free-standing house at the Georgia Institute of Technology constructed
to test new technologies. First, the robot autonomously learned to
operate 6 devices. After learning, we tested the robot's performance
in trials with each of the 6 devices for a total of 110 trials (110
trials $=$ (5 devices $*$ 2 behaviors $*$ 10 trials) $+$ (1 device $*$
2 behaviors $*$ 5 trials)). In all 110 trials, the robot autonomously
operated the device successfully after at most two attempts. If the
first attempt failed, the robot autonomously detected the failure and
then retrained using this new negative example prior to autonomously
trying a second time. We tested opening and closing drawers, turning
on and off light switches, and turning on and off rocker
switches. Figure \ref{fig_teaser} shows example output from the
resulting trained classifiers, which classify image feature vectors
as being associated with success or failure of a behavior.

\section{Related Work}

There is a significant body of work on robots learning to manipulate
autonomously, robots learning to perceive, perception for manipulation
that exploits task structure, and active learning methods for using
labeling efforts efficiently.  In this section, we discuss the
relationship between our work and current learning methods as well as
work that has demonstrated the effectiveness of using task-relevant
cues for perception in human environments. The research we present in
this paper builds on our earlier workshop publication
\cite{nguyen2011}.

\subsection{Robot Learning}

Even though the use of learning-based methods can yield powerful
detectors, labeled training examples are often time consuming and
expensive to obtain.  Different robot learning methods such as
imitation learning, interactive learning and developmental learning
\cite{lungarella2003a, pfeifer1997} can be grouped by how they
approach the issue of gathering data.  We now discuss different forms
of robot learning methods and specifically work that involves
autonomous learning, where the robot learns with little or no human
input after an initialization or teaching period.

\subsubsection{Autonomously Learning to Act}

Developmental learning research primarily uses data from the robot's
own interactions with its environment and emphasizes scenarios
inspired by the development of biological organisms
\cite{lungarella2003}.  Examples include studies of gaze control
\cite{berthouze1997, berthouze1998, butko2010}, reaching
\cite{metta1999, hulse2011}, pointing \cite{marjanovic1996}, and
poking \cite{metta2003}.

In the manipulation literature there has also been interest in using
autonomous learning based methods to find stable grasps.  One of the
earliest investigations in grasp learning is by Dunn and Segen
\cite{dunn88}) that matched objects using visual features and learned
candidate grasps through trial and error.  In \cite{Zhang2004},
instead of learning one grasping classifier the system learned
separate classifiers for grasp position and grasp orientation.  Saxena
et al.  \cite{Saxena2008} presented a method that learned a grasp
point classifier using a data set of simulated grasps and was able to
show success in grasping real objects in uncluttered environments.
Similarly, researchers in \cite{Erkan2010} also showed a grasp
learning algorithm, mapping from 3D edge features using an active
learning approach combined with semi-supervised learning.  The authors
of \cite{Klingbeil2008} used a supervised classifier to detect handles
and distinguish between left and right facing versions.  Work in
\cite{Montesano2009} from the developmental learning community views
the same problem as one of learning object affordances and proposed a
method for estimating grasp densities in images of objects on plain
backgrounds.

Research on learning from demonstration has investigated how to learn
policies and controllers for manipulation but, in most cases, has not
addressed perceptual challenges.  We refer readers to \cite{Argall08}
for an extensive survey of existing methods.

While many projects focus on inferring policies from human
demonstrations \cite{Argall08}, there is a subset of work where
autonomous learning is used with dynamic motion primitives
\cite{ijspeert03} to refine initial human demonstrations.  Using this
framework, \cite{Kober2010} present methods for parameterizing motions
refined from human demonstrations based on task objectives.  More
recently, Pastor et al.  \cite{Pastor2011} implemented this framework
on the PR2 robot to show the PR2 flipping a box using chopsticks and
playing pool.  In contrast, work in the domain of helicopter acrobatic
flights \cite{tang10} used a combination of human demonstrations and
approximate models, instead of real world practice, to extract
intended trajectories from sets of noisy demonstrations.  In the
system presented by by Prats et al.  \cite{prats12}, the authors
eschewed trajectories and instead focused on creating a system that
records task properties in terms of forces applied with respect to a
user-defined visual reference frame.

Using a generative method with a planning framework, Stulp et
al. \cite{stulp12} introduced the concept of action-related places,
modeling how a robot's navigational uncertainty affects its ability to
execute grasps.

\subsubsection{Autonomously Learning to Perceive}

In contrast to motor learning, most work in learning for perception
relies on data captured manually \cite{ponce2006}, captured in
simulation \cite{Klingbeil2008, Saxena2008}, or downloaded from the
web \cite{semantic_robot, chatzilari2011}.  Although large data sets
can be collected from these sources, the data generated can be biased
and may not match what the robot will encounter. Likewise, the
relationship between these data and the robot's actions may not be
clear. For example, a good location for a person to grasp or a
location that a person believes would be good for grasping may not be
appropriate for a robot, given its end effector and other
differences. Additionally, there are challenges in processing and
generating such data.  Text on the web is written in natural language
for human beings \cite{tenorth11}. Well-framed aesthetically pleasing
images can be misleading for robots using noisy sensors from
opportunistic points of view.  Accurate simulation of physical objects
can also be hard to obtain \cite{abbeel06b}.

The system that we present uses data generated from self-experience,
(similar to \cite{salganicoff1996, Erkan2010, Pastor2011}, and
\cite{Kober2010}).  With self-generated data, generalization becomes
less of an issue as the training data and data encountered online are
sampled from the same distribution.  Even so, labeled examples can be
costly to obtain.  Interactions with the environment take time, and
can potentially result in damage to the environment and the robot.
Human labeling can be labor intensive and have errors, ambiguity, and
inconsistencies \cite{barriuso12}.  We address this issue in our work
by combining active learning, which reduces the number of examples
needed, with autonomous learning methods that eliminate the need for
human labeling beyond an initialization process.

Past work in learning for perceptual categorization, a process where
agents learn through interaction with the world to divide sensory
information into distinct groupings, has used robot self-generated
data.  However, most systems were designed to classify only simple
geometric objects such as cylinders and rectangles using cross-modal
information \cite{krichmar2002, coelho2001, scheier1996}.

A relatively small subset of work investigates more complex objects
found in human environments.  In \cite{stober2011}, Stober et
al. demonstrated an approach for extracting spatial and geometric
information from raw sensorimotor data.  Sukhoy and Stoytchev
\cite{Sukhoy2010} presented an active learning method for a robot
pressing doorbell buttons.  Kraft et al. \cite{kraft2010} presented a
system that gradually learns object representations and associates
them with object-specific grasps.  Katz and Brock in \cite{Katz2008a}
showed a method with which a robot determines the structure of
articulated objects through experimentation.  Similarly, Hoof et
al. \cite{hoof12} presented a system that selects maximally
informative actions to segment tabletop scenes.

Paolini et al. in \cite{paolini12} present related work in which a
stationary robot estimates which parameters to provide to a robot
behavior based on experience. Like our approach, the parameters to the
behavior are in terms of a location. For them, the parameters to the
behavior describe the pose of a highlighter held in the robot's
gripper and the behavior attempts to stand the highlighter
upright. Their approach uses estimated probability distributions,
including a distribution that gives the probability of the marker's
pose conditioned on haptic sensing. In contrast, we use a
discriminative approach and directly map visual sensing to 3D
locations that are likely to result in success of the behavior.

The work of \cite{Sukhoy2010} is notable for its similarity to our
approach. They presented a system that uses an uncertainty sampling
scheme to actively learn the appearance of doorbell buttons.  In
contrast, our approach uses a different active learning algorithm,
works with a mobile manipulator operating in situ devices, and handles
persistent change to the state of the world.

\subsection{Task-Relevant Feature Detection}

In a parallel thread to robot learning, there has been recognition in
the mobile manipulation community of the importance of exploiting task
structure to reduce the complexity of operating in the real-world.
This point was argued by Katz et al. in \cite{Katz2008}.  Dang and
Allen \cite{Dang2010} showed evidence that many manipulation tasks can
be described using sequences of rotations and translations.
Additionally, work in articulated object perception \cite{Katz2008a},
tool tip detection \cite{Kemp2006}, door handle detection
\cite{Klingbeil2008}, behavior-based grasping \cite{Jain2009}, and
corner detection for towel folding \cite{Maitin-shepard2010} suggests
that, in many tasks, recovery of complex representations of the state
of objects prior to manipulation is unnecessary.  For example, Jain
and Kemp \cite{advait10} demonstrated that overhead grasping of
diverse real-world objects can be successfully accomplished by
representing a segmented object as a planar ellipse.  In addition to
detecting features used to parameterize manipulation behaviors, task
specific cues can be employed to verify the effects of a robot's
actions.  For example, Okada et al. have demonstrated the value of
task-specific perception for success detection by humanoid robots,
including detecting that liquid is flowing \cite{okada06}. These
systems illustrate that low-dimensional task-specific object
representations can result in good performance over real-world
variation. However, they used hand-coded and hand-trained feature
detectors that required significant engineering effort. With our
approach, robots autonomously learn to classify visual features as
being relevant to the success of a specific behavior or not.

\subsection{Active Learning and Curiosity Driven Learning}

In many robot learning scenarios, unlabeled data can be readily
acquired but labeling the data is costly. Researchers have proposed
the use of active learning methods \cite{settles12} to gain more value
from limited labeling. In many active learning algorithms, at each
iterative learning step the learner is given an option to select a
data point to be labeled out of a set of unlabeled data points.  For
one class of proposed approaches, the learner picks the data point
whose label it is most uncertain about \cite{lewis94, culotta05,
  settles08}.  With disagreement-based methods, learner ensembles
select the data point they most disagree on \cite{cohn94}. More
computationally demanding methods, however, attempt to explicitly
minimize future expected error or variance \cite{roy01, berger96,
  lafferty01}.  There are also proposals to combine semi-supervised
and active learning to exploit structure in unlabeled data
\cite{mccallum98, muslea02, zhu03}.  Although there have been several
large scale studies of active learning methods on different data sets
showing its superiority over randomly picking data points for labeling
\cite{korner06,schein07,settles08}, the best active learning algorithm
to use in each circumstance is application specific.  In our work, we
use a heuristic that picks the data point closest to the decision
boundary of a support vector machine (SVM) for labeling, a method that
has been shown to perform well in a variety of applications
\cite{jain10b, Schohn2000, tong00}.

\section{Approach}

Our approach enables a mobile manipulator to autonomously learn a
function that takes a 2D RGB image and a registered 3D point cloud as
input and returns a 3D location at which a manipulation behavior is
likely to succeed. Our approach requires a pair of manipulation
behaviors, verification functions that detect when each behavior has
been successful, and an initial hint as to where one of the behaviors
will be successful.

Each behavior must have input parameters that correspond with a 3D
location that specifies where the behavior will act. During training,
our system executes each behavior multiple times using different 3D
locations around the device being manipulated and records whether or
not the behavior succeeded at each location. For each 3D location, the
system creates an image feature vector using an area of the registered
2D RGB image associated with the 3D location. These image feature
vectors are labeled with whether or not the behavior succeeded or
failed at their associated 3D locations. In other words, the collected
data set consists of positive and negative examples of image feature
vectors that were or were not associated with the success of the
behavior.  With a classifier trained from this data set, the robot can
then predict if the associated behavior will succeed at a 3D location
based on the image feature vector associated with the location.

To avoid user intervention during training, our procedure trains two
behaviors at the same time, switching to the other behavior when the
current behavior succeeds.  This enables our method to operate devices
that can be approximated as having two binary states, such as a drawer
being open or closed.  Using a pair of behaviors allows the robot to
change the device back and forth between these two states, so that
training can continue autonomously. For example, instead of training a
drawer opening behavior in isolation, our process flips to training a
drawer closing behavior when opening succeeds and vice versa until the
classifier converges. We also formalize the relationship between the
two behaviors and define them as complementary behaviors.

Using self-generated data takes considerable time, since each labeled
image feature vector requires that the robot execute the behavior at a
3D location and observe the results. To avoid needing an intractable
number of trials, our method uses active learning to execute the
behavior at an informative 3D location at each iteration.
Specifically, our procedure trains a support vector machine (SVM)
after each trial using the current labeled data. It then uses a
heuristic proposed by Shohn and Cohn \cite{Schohn2000} to select the
unlabeled image feature vector that is closest to the current SVM's
decision boundary to be labeled next. It then executes the behavior at
the 3D location associated with this image feature vector.

Our training procedure has two phases. The first is an initialization
phase where the user selects the behavior pair to train, gives a seed
3D location, and positions the robot's mobile base for training. The
next phase is an autonomous phase where the SVM active learning
procedure runs until the learner converges.  After convergence, each
behavior has a classifier that predicts 3D locations where it will
succeed.

During runtime, if the behavior's verification function detects a
failed attempt, our procedure appends this negative example to the
data set, retrains the classifier, and tries again using the output of
this new classifier (Section \ref{sec_retry}).

In the following sections, we discuss the requirements of our learning
procedure \ref{requirements_sec}, properties of complementary
behaviors (Section \ref{complementary_behaviors_sec}), our training
procedure in detail (Section \ref{autonomous_training_sec}), and
classification infrastructure (Section \ref{classification_sec}).

\subsection{Requirements}
\label{requirements_sec}

For our algorithm to apply, several assumptions must be met. First,
our approach assumes that the robot can execute a set of behaviors,
each of which only requires a 3D location in the robot's frame of
reference as initial input.  We have demonstrated that this is a
reasonable assumption for a variety of useful mobile manipulation
behaviors in our previous work on laser pointer interfaces
\cite{nguyen08}.

Second, this approach assumes that the robot has a way of reliably
detecting whether or not a behavior it has executed was successful or
not.  Our approach assumes that the verification function, $V$,
returns whether or not a behavior succeeded. For this work, it takes
the form $V(I(b), I(a))$, where $I(x)$ is the array of robot sensor
readings when the state of the world is $x$.  The states $b$ and $a$
are the states before and after the robot executes a behavior.

Third, the approach assumes that for each behavior, $B$, there is a
complementary behavior, $B^{*}$.  If $B$ successfully executes, then
successful execution of $B^{*}$ will return the world to a state that
allows $B$ to execute again.  We discuss the implications of this
requirement in Section \ref{complementary_behaviors_sec}.

Fourth, each behavior must return a 3D location indicating an
approximate location where its complementary behavior should execute.
This requirement is provided as a device for the behavior pair to
communicate the position of the object manipulated to each other.
These requirements can be summarized as:

\begin{align}
\label{eqn_behavior}
behavior(p_{3D})     \mapsto (success, p_{3D}) \\
behavior^{*}(p_{3D}) \mapsto (success, p_{3D})
\end{align}

\subsection{Complementary Behaviors}
\label{complementary_behaviors_sec}

\begin{figure}
\centering
\includegraphics[height=8.0cm]{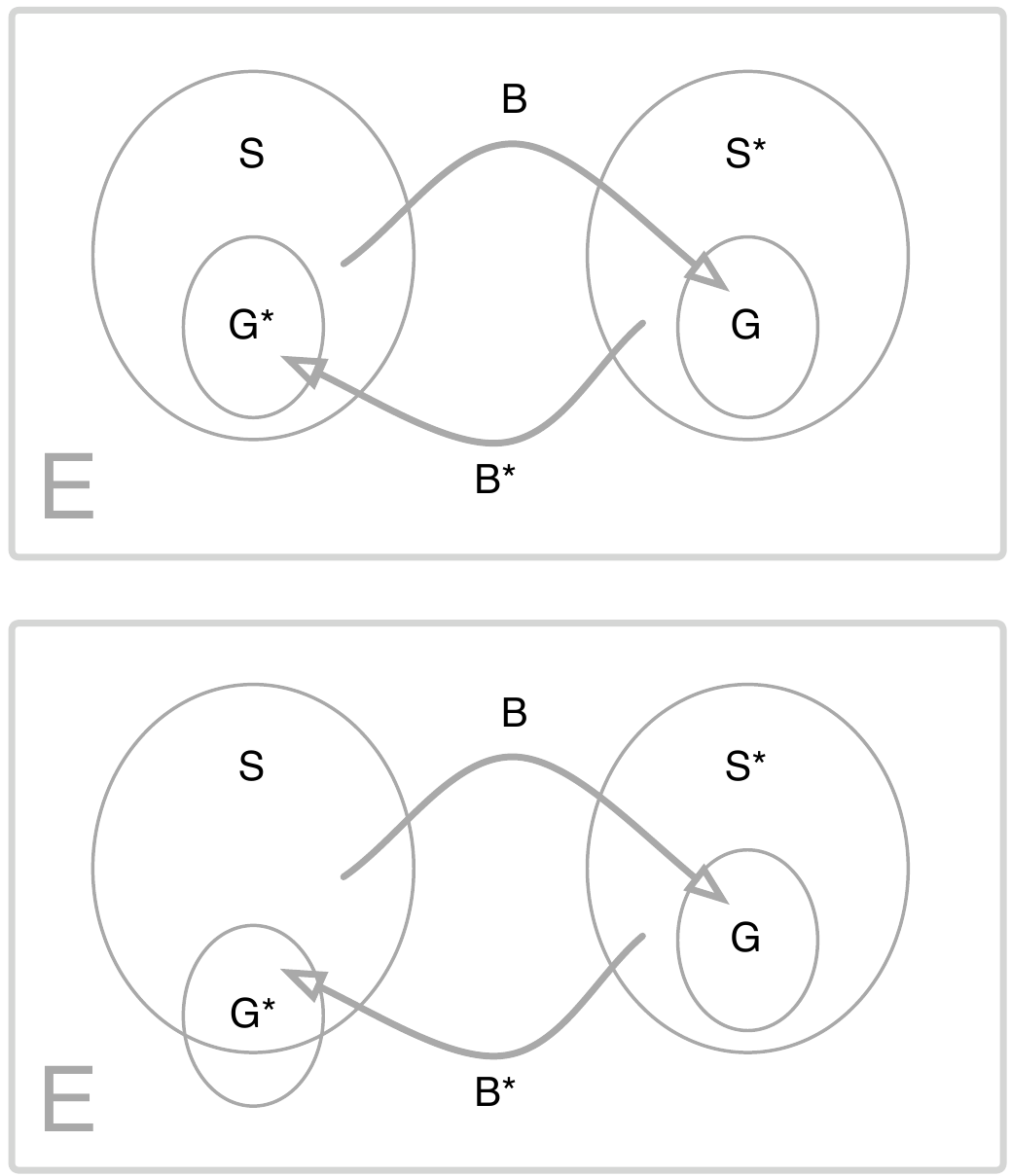}
\caption{ Relationships between set $S$, $G$, $S^*$, $G^*$, $B$, and
  $B^*$.  \textbf{Top:} An example set of complementary behaviors
  where $G^* \subseteq S$ and $G \subseteq S^*$.  In this case, the
  effect of $B$ is reversible using $B^*$.  \textbf{Bottom:} An
  example set of behaviors that are not complements with $G^*
  \nsubseteq S$, so $B^*$ can produce states that are not in $S$.
\label{fig_complement_non_complement}
}
\end{figure}

In order to train autonomously, without human intervention, our
procedure uses a complementary pair of behaviors during its data
gathering process.  We introduce the notion of a complementary robot
behavior $B^*$ to a behavior $B$ as being a behavior that is capable
of ``reversing" the state of the world, so that behavior $B$ can be
used again.  For example, if behavior $B$'s function is to turn off
the lights using a light switch, its complement, $B^*$, would turn the
lights back on using that light switch.  If a behavior opens a door,
then its complement would close the door.

We formalize our notion of complementary behaviors by defining the
relationship between ideal complementary behaviors. We first define a
hypothetical state space $E$ that contains the states of
everything in the world, including the robot's state.  We
then represent execution of behavior $B$ given an initial state of the
world $i \in E$ as $B(i)$, where $B$ is an operator that takes the
initial state of the world $i$ as input and returns the resulting
state of the world $r \in E$. Furthermore, when $B$ is applied to a
state $s \in S$, where $S \subseteq E$ is a set of \textbf{s}tarting
states, it returns $g \in G$, where $G \subseteq E$ is a set of
goal states. We define \begin{equation} G = \{ g | V(I(i),
  I(g)) = success \wedge g = B(i) \wedge i \in E \}
\end{equation} and \begin{equation}
S = \{ s | g \in G \wedge g = B(s) \wedge s \in E \}.
\end{equation} Intuitively, if the state of the world, $s$, is a start state, $s \in
S$, then the behavior $B$ will be successful and the resulting state
of the world, $g = B(s)$, will be a goal state, $g \in G$.

We now define a complement $B^*$ of behavior $B$ to have a set of
start states, $S^*$, and a set of goal states, $G^*$, such that $G^*
\subseteq S$ and $G \subseteq S^*$ (see Figure
\ref{fig_complement_non_complement}).  This guarantees that applying
$B$'s complement, $B^*$, after successfully applying $B$ will result
in a state of the world that allows $B$ to once again be applied
successfully. More formally, it guarantees that $B^*(B(i)) \in S$ when
$i \in S$, and that $B(B^*(i)) \in S^*$ when $i \in S^*$.

\subsection{Autonomous Training}
\label{autonomous_training_sec}

\begin{figure} [t]
\centering
\includegraphics[height=4.0cm]{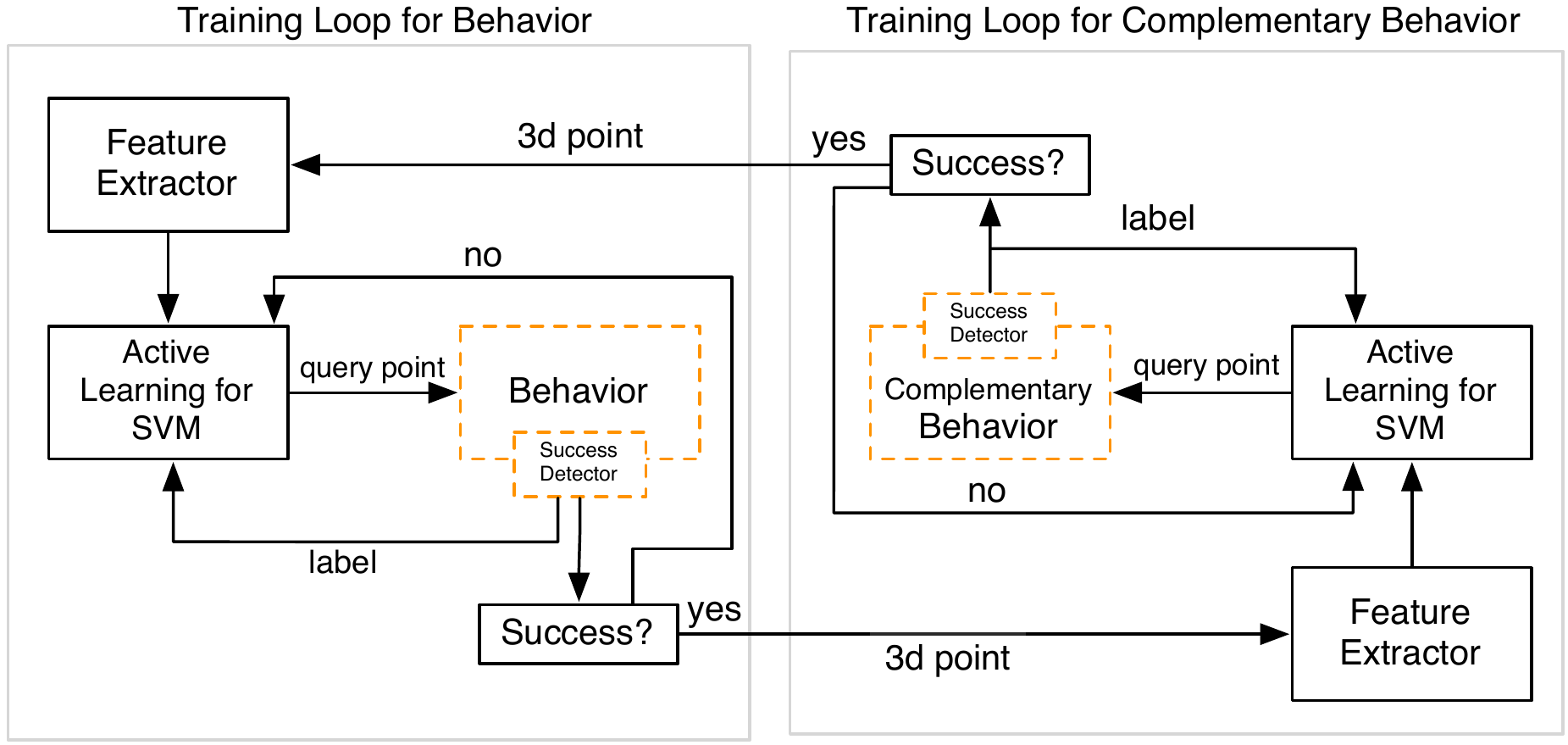}
\caption{\label{fig_learning_flip_flop} Illustration of the classifier
  training procedure where the system trains the complementary
  behavior upon success of the first behavior and vice versa.  Dashed
  orange boxes on the two behaviors and success detectors highlight
  that these modules are provided as input to our system. }
\end{figure}

\subsubsection{Initialization} 
\label{sec_initialization}

\begin{figure*}[th]
\centering
\includegraphics[height=5.0cm]{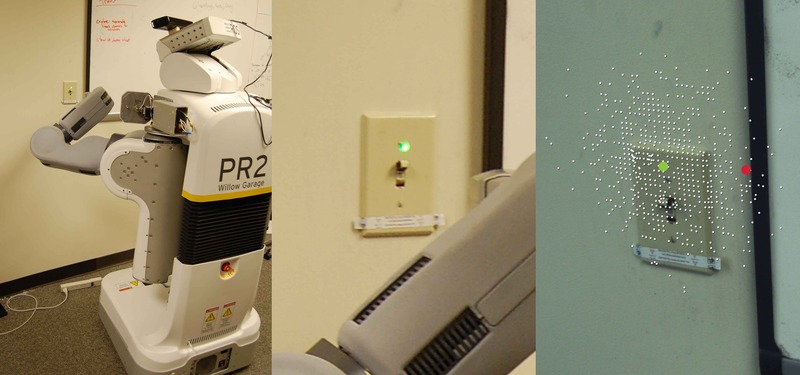}
\caption{Illustration of the initialization procedure for a pair of
  behaviors that flip light switches. \textbf{Left} Position robot in
  front of the switch.  \textbf{Middle} Illuminate an initial 3D
  location as input to the behavior using a laser pointer.
  \textbf{Right} A 3D location associated with success (green) and a
  3D location associated with failure (red) after
  initialization. Unexplored 3D locations are shown in white.}
\label{initialization_fig}
\end{figure*}

Our initialization procedure is motivated by the scenario in which a
user would take the robot on a home tour and point out 3D locations
using a green laser pointer \cite{nguyen08} and specify behaviors
applicable to those locations.  After this tour, the robot would later
autonomously navigate back and learn to robustly perform the
behaviors.

For this paper, we have implemented an initialization procedure that
starts with the user navigating the robot to be in front of the device
to be operated using a gamepad interface. Then using a green laser
pointer \cite{nguyen08}, the user designates an initial 3D location to
begin exploring. The robot samples 3D points around this designated
location (using a spherical Gaussian with a variance of 4 cm) and
executes the behavior pair with respect to them.  After each execution
of a behavior at a 3D location, the behavior's verification function
returns a label of either success or failure.  The sampling process
continues until the procedure gathers data points from at least one
successful and one failed trial.  These two data points are then used
to train SVM classifiers that guide the data gathering process with
the active learning heuristic \cite{Schohn2000}.

After this initialization, the robot stores a 2D mobile base pose
with respect to a global map, the user provided 3D location, an SVM
trained using two labeled data points, and labels indicating which
pair of behaviors is applicable at the specified location. We
illustrate this procedure in Figure \ref{initialization_fig}. In
addition, the user navigates the robot to eight different poses in
the room, referred to as practice poses, each at least a half
meter away from the device. The robot also stores the 2D mobile base
poses associated with these eight practice poses.

\subsubsection{Training Procedure}
\label{sec_train}

\begin{figure} [t]
\centering
\includegraphics[height=3.0cm]{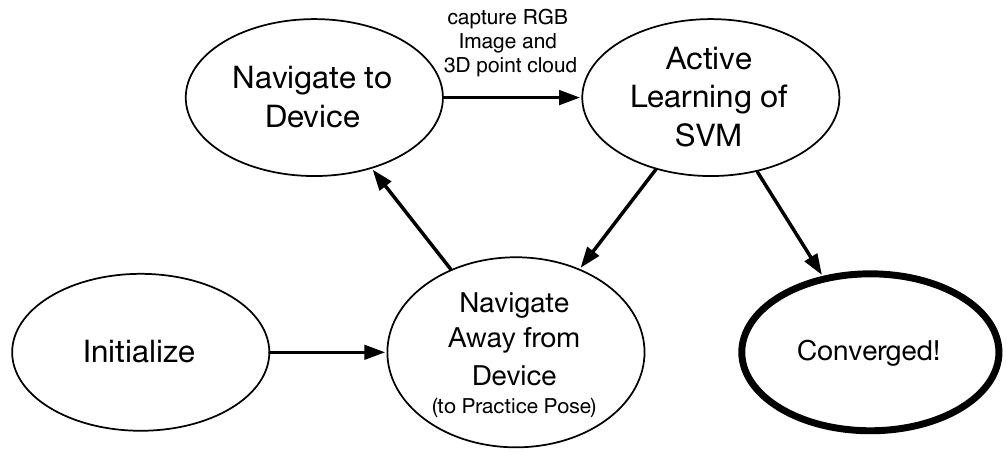}
\caption{Overview of the training procedure: initialization of the
  classifier; specification of practice poses in the environment by
  the user; and a loop that navigates the robot to each practice pose
  and back to the device until the robot gathers enough training data.
\label{fig_learning_behavior}}
\end{figure}

\begin{figure} [t]
\centering
\includegraphics[height=5.0cm]{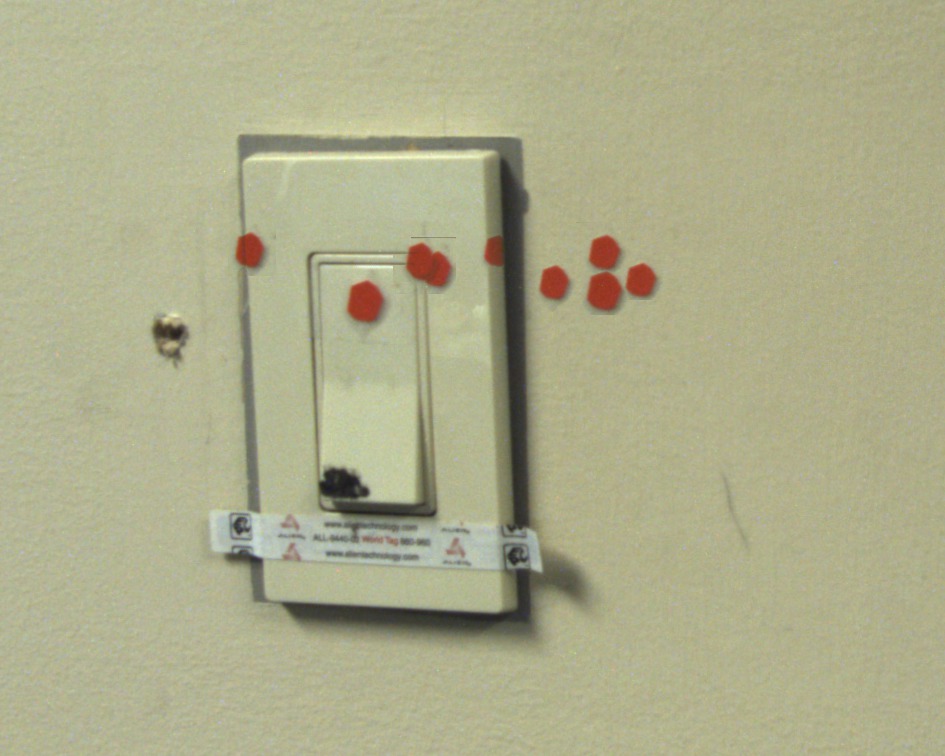}
\caption{This figure shows a visualization of task variation due to
  the robot's mobility. We affixed a red dot at the center of a rocker
  switch. The robot attempted to navigate to the same pose and
  take the same picture of the switch 10 times. This image
  superimposes the red dot from 9 images onto the first image to
  illustrate the wide variation due to navigation. One of the 10 dots
  is obscured by 2 others. The switch plate shown has a width of 7.0
  cm. If the robot were to use its localization estimate to press this
  switch, most of the attempts would result in failure.
\label{fig_open_loop}}
\end{figure}

Our training procedure is designed to emulate conditions that the
robot would encounter when performing the task. After receiving a
command, the robot navigates to the device so that it can execute the
commanded behavior.  Navigation and localization errors result in
variations that can substantially reduce the performance of a
behavior, such as variation in the robot's point of view.  We
illustrate task variation due to navigation in Figure
\ref{fig_open_loop}. Our training method samples from this source of
task variation by commanding the robot to navigate to one of eight
practice poses in the room and then commanding it to navigate back
to the device (see Figure \ref{fig_learning_behavior}).

After navigating to the device, our procedure begins an active
learning phase (see Figure \ref{fig_learning_flip_flop}).  We
summarize this phase in Algorithm \ref{practice_alg}. The process
starts with the robot capturing an RGB image and a registered 3D point
cloud. The robot then computes image feature vectors for 3D points
randomly sampled from the point cloud around the device
(\emph{extract\_features}). It then iteratively selects image feature
vectors (\emph{svm\_pick}) that it labels by executing the behavior at
the associated 3D location and using the verification function
(\emph{execute\_behavior}).  After each trial, the process retrains
the SVM classifier with a data set that incorporates the newly
acquired example (\emph{add\_instance\_and\_retrain\_svm}).  The
procedure stops after gathering a maximum of six labeled image feature
vectors or the learner converges (\emph{stop\_criteria}).  We imposed
this conservative maximum limit because image feature vectors gathered
from the same view are correlated, which can confuse the learning
heuristic and result in the training process stopping prematurely. If
the robot operates the device successfully in a trial, the algorithm
continues, but uses the complementary behavior for the next iteration
(section following \emph{If(success)}).

\begin{algorithm}[t]
    instances, $candidates^{3D}$ = extract\_features($point^{3D}$);

\While{True}
{
    instance, $candidate^{3D}$ = svm\_pick(behavior, instances, $candidates^{3D}$);

    \If{stop\_criteria(behavior) or svm\_converged(behavior, instances)}
    {
        break;
    }

    success, $candidate^{3D*}$ = execute\_behavior(behavior, $candidate^{3D}$);

    add\_instance\_and\_retrain\_svm(instance, success);

    $instances = instances \setminus instance$;

    $candidates^{3D} = candidates^{3D} \setminus candidate^{3D}$;

    \If{success}
    {
        practice($candidate^{3D*}$, comp\_behavior, None,
        stop\_criteria=stop\_on\_first\_success);
    }
}
\caption{practice($point^{3D}$, behavior, comp\_behavior, stop\_criteria)}
\label{practice_alg}
\end{algorithm}

This process continues until \emph{svm\_converge} is satisfied for
each of the eight practice poses. Once it is satisfied for a
particular practice pose, the robot no longer navigates to the
pose. We define convergence for a practice pose to occur when
after driving up to the device from the practice pose, none of the
initially computed image feature vectors are closer to the decision
boundary than the current support vectors.

\subsubsection{Behavior Execution Procedure}
\label{sec_retry}

The training process above produces a classifier that can reliably
detect locations where the associated behavior will succeed.  To use
this classifier, our robot navigates to the device using the 2D map
pose stored during initialization, classifies 3D points in the view
that it sees, finds the mode of the positive classified points using
kernel density estimation \cite{scipy_kde}, selects the 3D point in the point cloud
closest to this mode, and executes the associated behavior using the
resulting 3D location.

If the behavior fails to execute using this 3D location, our procedure
adds the associated image feature vector as a negative example to the
data set and retrains the classifier.  This new example changes the
classifier's decision boundary. The robot then selects a new 3D
location using the retrained classifier with the originally computed
image feature vectors. This continues until the behavior is
successful. It then adds the image feature vector associated with this
success to the data set as a positive example and retrains the SVM. In
contrast to systems where the execution process is independent of data
gathering and training, the robot has the opportunity to retrain its
classifier when it detects errors made during execution, giving the
possibility of lifelong training.

\subsection{Classification}
\label{classification_sec}

The base classifier that we use in this work is a support vector
machine (SVM).  As is standard in supervised classification, given a
data set of labeled examples $D = \{(x_1, y_1), ... $ $(x_N, y_N)\}$
with $x_i \in \mathbb{R}^M$ representing feature vector i of local 2D
appearance information associated with a candidate 3D point and $y_i
\in \{1, -1\}$ where positive and negative denote, respectively,
success and failure, we want to be able to predict $y_j$ for a future
instance $(x_j,y_j) \notin D$.

As functional structures on many household devices are often small
compared to nonfunctional components, such as the size of a switch
relative to the plate or wall, there is typically an unbalanced data
set problem, since there can be many more negative than positive
examples.  In unbalanced data sets the SVM can return trivial
solutions that misclassify all the positive samples, since the
misclassification cost term in the SVM objective is defined over all
samples.  To prevent this issue, we use an SVM formulation that
separates the costs of misclassifying the negative class from the cost
of misclassifying the positive class \cite{Chang2011}, \begin{eqnarray}
\min_{w, b, \xi} && \frac{1}{2}\mathbf{w}^T\mathbf{w} + C^+\sum_{y_i=1}\xi_i + C^-\sum_{y_i=-1}\xi_i \nonumber \\
s.t.             && y_i(\mathbf{w}^T\phi(\mathbf{x}_i)+b) \geq 1-\xi_i  \label{eq:svm_balanced} \nonumber \\
                 && \xi_i\geq 0, i=1,\ldots, l  ,
\end{eqnarray} where $\mathbf{w}$ and b are SVM parameters, $\xi_i$ counts the margin
violations for misclassified points (in the case of nonseparable
data), and $\phi()$ is the radial basis kernel function we use
(discussed in Section \ref{sec_parameters}).

This formulation separates the SVM misclassification cost scalar $C$
into $C^+$ and $C^-$ which are, respectively, costs due to negative
and positive misclassifications. For our system, we set $C^-$ to be 1,
and $C^+$ to be the number of negative examples over the number of
positive examples.  This scaling keeps the percentage of misclassified
positive and negative examples similar in our skewed data set, where
there might be many more negative than positive examples.  Without
this adjustment, we found that training often returned trivial
classifiers that classified any input vector as negative.

\subsubsection{Active Learning Heuristic}

Our training process iteratively builds a data set that it uses to
train the classifier. Before each trial, the system selects the image
feature vector to label. To select the feature vector, the system uses
a heuristic developed in \cite{Schohn2000} that selects the feature
vector closest to the decision boundary of the existing SVM, under the
condition that it is closer to the boundary than the SVM's support
vectors. The procedure converges when no feature vectors remain that
are closer to the decision boundary than the support vectors.

At each iteration $i$ of our procedure, we define the previous
iteration's data set as $D_{i-1}$, the current set of support vectors
as $X_{i}^{sv}=\{x_1^{sv}, \ldots, x_P^{sv}\}$, the unlabeled image
feature vectors as $X_i^{q}=\{x_1^q, \ldots, x_M^q\}$, and the SVM
distance function, which measures distance to the decision boundary,
as $d(\mathbf{x_i})=$ $\left | \mathbf{w}^T\phi(\mathbf{x}_i)+b \right
|$. The system selects the unlabeled image feature vector that is
closest to the decision boundary as specified by the following
expression: \begin{equation} \underset{\mathbf{x}_i^q: \forall
    \mathbf{x}_j^{sv}\;d(\mathbf{x}_i^q) <
    d(\mathbf{x}_j^{sv})}{\operatorname{argmin}} \quad
  d(\mathbf{x}_i^q)
\end{equation}

\subsubsection{Features}
\label{sec_features}

\begin{figure*} [t]
\centering
\includegraphics[height=7.5cm]{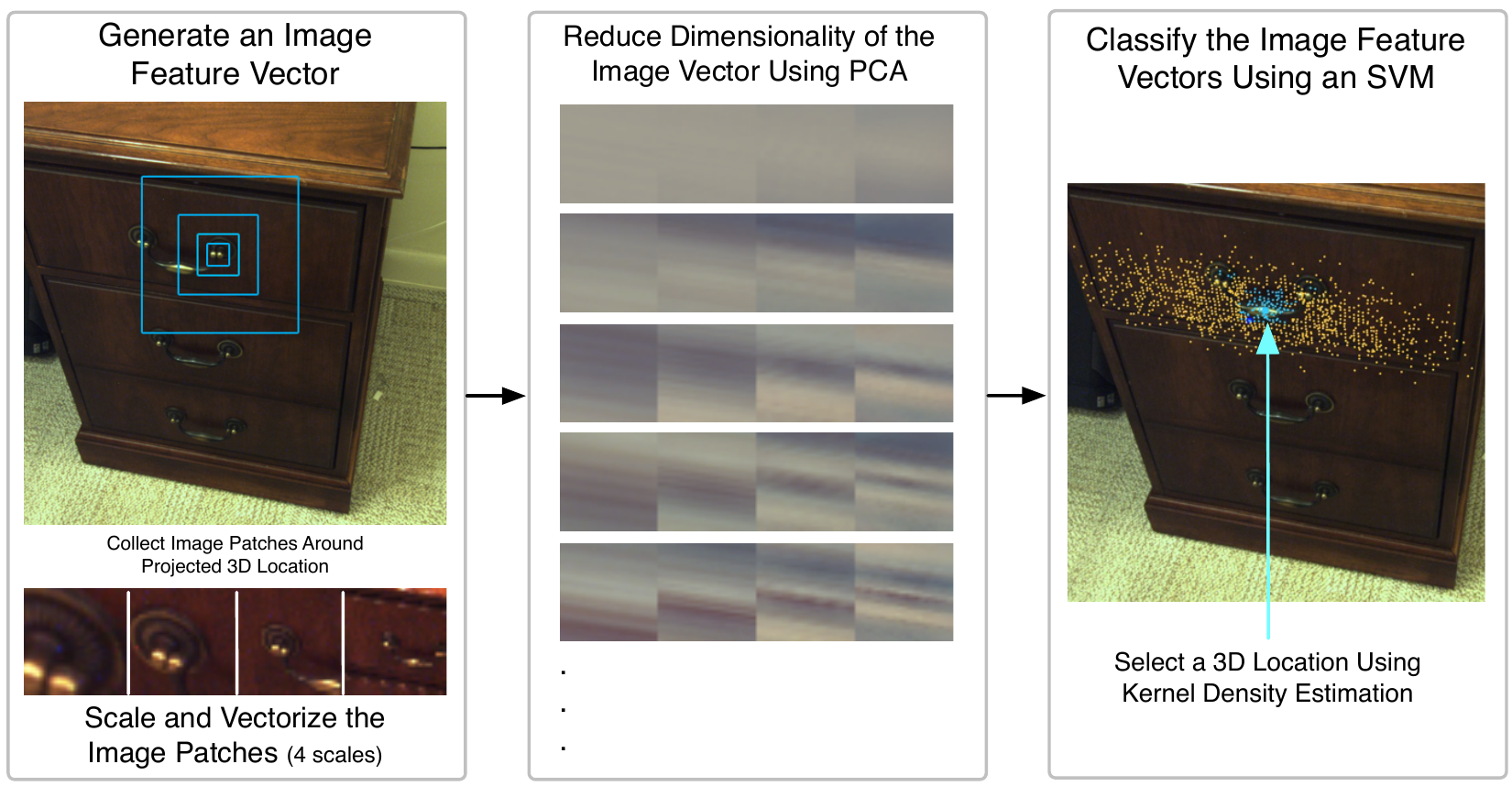}
\caption{\label{fig_classification_diagram} To select a 3D location at
  which the behavior is likely to be successful, the system first
  generates image feature vectors for a set of 3D locations. It does
  so by vectorizing and then reducing the dimensionality of scaled
  image patches centered around the 2D projection of each 3D
  location. Then it uses an autonomously trained SVM to classify each
  of these image feature vectors as predicting success (blue) or
  failure (orange) of the behavior. Finally, it selects a specific 3D
  location using kernel density estimation.}
\end{figure*}

The feature generation procedure, which is illustrated in Figure
\ref{fig_classification_diagram}, takes as input a 3D point cloud, a
registered high resolution RGB image, and a reference 3D point.  The
system first selects random 3D points from the point cloud, without
replacement, around the reference 3D point according to a Gaussian
distribution $\mathcal{N}(\bar{p},\, \Sigma)$, where $\Sigma =
diag(v_x, v_y, v_z)$ with $v_x, v_y,$ and $v_z$ being, respectively,
variances in the x, y, and z direction.  The Gaussian mean
$\bar{\mathbf{p}}$ is set to the 3D reference point.  This Gaussian
search prior enables the system to save computational effort and focus
its attention on the device that the robot is supposed to manipulate.

After randomly selecting a set of 3D points, the system projects each
3D point $\mathbf{p}_i^c$ into the high resolution RGB image,
$proj(\mathbf{p}_i^c)$. For each projected 3D point, it collects
square image patches of successively increasing size centered at the
projected 2D point in the RGB image, scales these patches to have the
same height and width, vectorizes them, and concatenates them into an
image feature vector. The system then uses Principle Components
Analysis (PCA) to reduce the dimensionality of these image feature
vectors. We discuss the specifics of these steps in Section
\ref{sec_parameters}.

\section{Implementation}

\subsection{Learner Parameters}
\label{sec_parameters}

We implemented our system on on a PR2 robot \cite{willow_website}: a
mobile manipulator produced by Willow Garage with two arms, an
omnidirectional base, and a large suite of sensors. Our system uses 3D
point clouds and 5 megapixel RGB images from the robot's tilting laser
range finder and Prosilica camera.

Starting with a 3D point cloud and registered RGB image, our process
randomly selects 3D points from the point cloud as described in
Section \ref{sec_features}.  For each selected 3D point, the system
collects image patches at 4 scales centered around the point's 2D
projection in the RGB image. The raw image patches have widths of 41,
81, 161, and 321 pixels. They are then scaled down to be 31x31 pixel
image patches, vectorized, and concatenated into an 11,532 element
image feature vector for each 3D point. The vectors are then reduced
to 50 element vectors by projecting them onto PCA basis vectors that
are calculated for each action using the 11,532 element image feature
vectors computed from the first 3D point cloud and RGB image captured
during initialization.

To classify these 50 dimensional image feature vectors, we use SVMs
with radial basis function kernels.  We set the hyperparameters of
this kernel using an artificially labeled data set.  To create the
data set we took 10 different 3D point clouds and RGB images of a
light switch from different views and geometrically registered them.
After hand-labeling one 3D point cloud and RGB image, we geometrically
propagated labels to the other 9.  To find the kernel hyperparameters,
we split the labeled image feature vectors from this data set into a
training set and a test set. We then performed a grid search
\cite{Chang2011} for the set of hyperparameters that best generalized
to unseen data in the test set.

\subsection{Behaviors}

To evaluate our system, we implemented three pairs of complementary
behaviors that operate light switches, rocker switches and
drawers. These tasks are sensitive to the location at which an action
is performed. For example, light switches are small targets that
require high precision and accuracy for the PR2 to operate with its
finger tips. As illustrated in Figure \ref{fig_open_loop}, we have
found that a PR2 will rarely succeed at flipping a light switch if it
simply navigates to a pre-recorded location and moves the arm through
a pre-recorded motion without visual feedback.

\subsection{Light Switch Behaviors}
\label{light_switch_behavior_sec}
\begin{figure} [t]
\centering
\includegraphics[height=5.5cm]{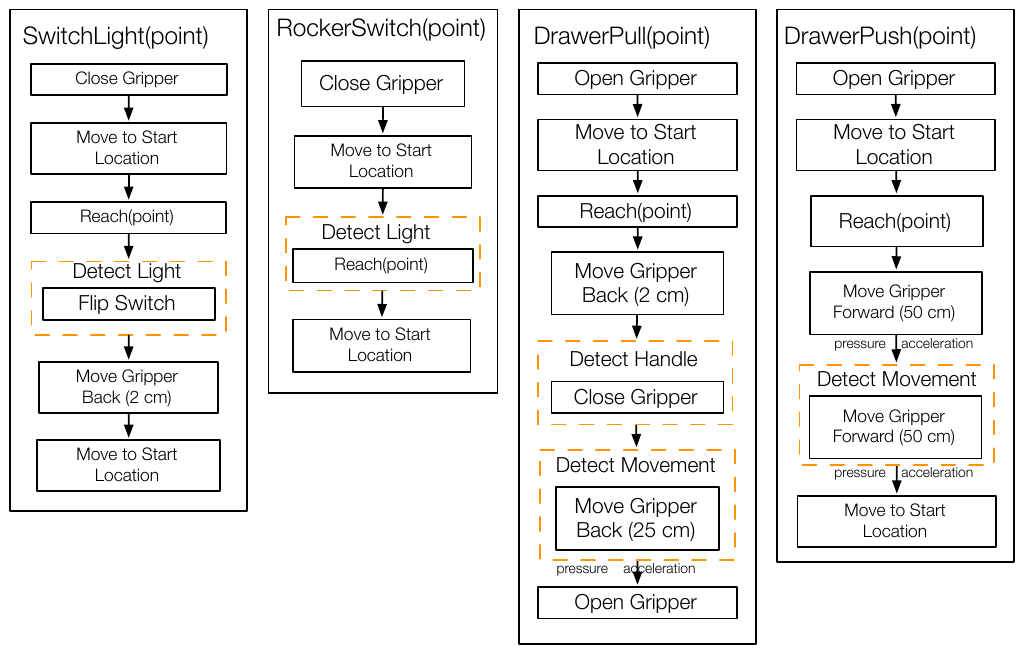}
\caption{\label{fig_mech_behaviors} Sequence of actions performed by
  each of the eight behaviors used in this work for operating a light
  switch, rocker switch and drawer. Dotted orange boxes indicate
  procedures for detecting success or failure in a given behavior.}
\end{figure}

Our light switch behavior's strategy is to reach forward to the
specified 3D location, stop on contact detected with gripper tip
tactile sensors, then slide along the contacted surface in the
direction of the switch.  A successful 3D location needs to place the
robot's finger so that its width will make contact with the switch and
far enough above or below the switch so that the finger will move the
switch down or up. Figure \ref{fig_mech_behaviors} shows the sequence
of actions taken by this behavior.

The behavior starts with the robot closing its gripper (Close
Gripper), moving the gripper to a pre\-/manipulation location (Move to
Start Location), reaching to the given 3D location (Reach), flipping
the switch by sliding along the flat surface (Flip Switch), moving the
gripper back (Move Gripper Back), then moving back to the initial
location (Move to Start Location).

There are a few steps in this behavior where the robot detects tactile
events. When reaching, the robot stops when it detects contact using
pressure sensors on its finger tips.  Next, the sliding movement stops
after detecting a spike in acceleration with the accelerometer
embedded in the robot's gripper.  In the context of this task, this
spike in acceleration typically corresponds with the light switch
flipping.

To detect success, our behavior measures the difference between the
average intensity of an image captured before sliding along the
surface and an image captured after. A large difference indicates that
the lighting intensity changed.

The complementary behavior is identical except for a change in the
direction of flipping.  After executing, the behavior and
complementary behavior return the 3D location input with a predefined
offset ($\pm$ 8 cm).

\subsection{Rocker Switch Behaviors}

Our rocker switch behavior consists solely of a reaching out step
similar to the light switch behavior above, since the force applied
from contact during the reach procedure is enough to activate the
switch.  A successful 3D location will result in the robot's fingers
pushing in the top or bottom of the rocker switch.

This behavior uses the same image differencing method to detect
success as the light switch behavior. It calculates the difference
between images captured before and after the robot reaches
forward. After executing, the behavior and complementary behavior
return the 3D location with a predefined offset ($\pm$ 5 cm).

\subsection{Drawer Behaviors}

Pulling open and pushing closed a drawer require different behaviors
and success detection methods.  Our pulling behavior reaches to the
drawer handle location, detects contact, moves back slightly, grasps
with the reactive grasper from \cite{hsiao2010}, and pulls. When
pulling, failure is detected if the grasp fails or the robot fails to
pull for at least 10 cm while in contact with the handle. A successful
3D location will result in the robot's gripper grasping the handle
well enough to pull it back by at least 10 cm. When pushing, failure
is detected if the gripper does not remain in contact with the surface
for at least 10 cm. This classifies events where the robot pushes
against a closed drawer or an immovable part of the environment as
failures. After executing, the behavior and complementary behavior
return the 3D location the tip of the gripper was in immediately after
pulling or pushing.

\section{Evaluation}

\begin{figure} [t]
\centering
\includegraphics[height=5.5cm]{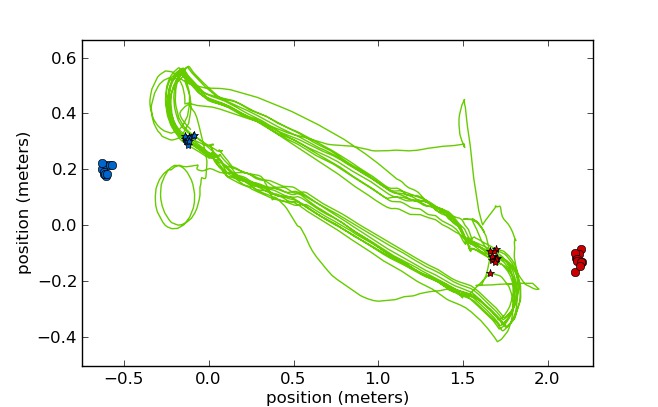}
\caption{\label{fig_positioning_errors} Results of experiments for
  which we used a motion capture system to track the robot's pose
  while navigating between two goal poses (blue and red).  Green is
  the path the robot took.  Stars indicate the final poses of the
  robot after it navigated to the goal poses. Circles show a point 50
  cm in front of the robot.}
\end{figure}

We evaluated our system using six separate devices.  We first tested
on a rocker switch using the PR2 robot named GATSBII in our lab, the
Healthcare Robotics Lab (HRL).  For the remaining five devices we
performed tests in the Georgia Tech Aware Home, a residential lab on
campus used as a test bed for new technologies.

In each environment, we began our evaluation by creating an occupancy
grid map of the area with the PR2's built-in navigation package
\cite{eitan2010}.  Then, after initialization (Section
\ref{sec_initialization}), we ran the autonomous training system
(Section \ref{sec_train}) until convergence. The experimenter provided
8 practice poses in the room representative of places from which the
robot might travel. The autonomous training system ran without
experimenter intervention except for pausing and resuming when the
robot's batteries ran low. In all, we trained 12 classifiers, a result
of having 6 devices and a pair of behaviors for each device (12 $=$ 6
$*$ 2).

After finishing the training sessions, we evaluated each classifier by
running each behavior multiple times, giving 110 trials in all (110
trials $=$ (5 devices $*$ 2 behaviors $*$ 10 trials) $+$ (1 device $*$
2 behaviors $*$ 5 trials)). During each trial we allowed the behavior
to retry and incorporate information from failures if it did not
succeed the first time.  However, we discarded any data gathered
during the retry procedure by previous trials at the start of each new
trial to obtain accurate error statistics for the original classifier.

For the devices we used in our evaluation, the functional components
are difficult for the PR2's laser range finder to detect.  Light
switches only show up as a few protruding 3D points similar to other
noisy 3D points produced by the sensor. The rocker switch appears as a
flat 2D texture on the 3D point cloud. Drawer handles tend to be
metallic and reflective resulting in an absence of 3D points. Using
features from RGB images enabled the robot to overcome these
challenges.

\subsubsection{Effects of Navigation Errors}

To better understand the variation in the task due to the robot's
mobility, we investigated how the pose of the PR2 varies when
navigating to a goal pose.  Using a room equipped with a NaturalPoint
OptiTrak motion capture system, we tracked the pose of the PR2 and
commanded the robot to navigate back and forth to two goal poses 10
times each.  As the standard deviation of the robot's Cartesian
position does not represent angular errors, we calculated errors for a
point 50 cm in front of the robot, which is representative of where a
device would be located.  The standard deviation of the location in
front of the robot was 1.85 cm, and 1.79 cm in the x and y directions,
respectively.  For the second position, the standard deviation was
1.55 cm and 2.38 cm in the x and y directions, respectively.  We show
the results of this experiment in Figure
\ref{fig_positioning_errors}. These errors demonstrate that navigating
to a pre-recorded location and moving the arm through a pre-recorded
motion would result in large variation that can result in failure. For
example, the robot's finger tips are 2.0 cm wide and light switches
are only 0.8 cm wide.

\subsection{Results}

\begin{table}
\centering
\caption{Training examples (abbreviated Ex.) gathered gathered for each action.
\label{table_training}
}
\begin{tabular}{cccc}
\hline
Action                & Positive Ex. & Negative Ex. & Total \\
\hline
HRL Rocker On         & 49           & 96           & 145 \\
HRL Rocker Off        & 47           & 94           & 141 \\

Aware H. Rocker On  & 26           & 47            & 73 \\
Aware H. Rocker Off & 29           & 52            & 81 \\

Ikea Drawer Open      & 23           & 35           & 58 \\
Ikea Drawer Close     & 23           & 39           & 62 \\

Brown Drawer Open     & 21           & 62           & 83 \\
Brown Drawer Close    & 25           & 46           & 71 \\

Orange Switch On      & 17           & 43           & 60 \\
Orange Switch Off     & 20           & 31           & 51 \\

Ornate Switch On      & 38           & 66           & 104 \\
Ornate Switch Off    & 40           & 76           & 116 \\
\hline
\end{tabular}

\end{table}

\begin{table}
\centering
\caption{For each trained behavior we ran 10 trials. We list the
number of tries until success for these trials below.  
\label{table_execution}
}
\begin{tabular}{ccc}
\hline
Action         & $1^{st}$ Try  & $2^{nd}$ Try  \\
\hline
HSI Rocker On         &  2            &  3            \\
HSI Rocker Off        &  4            &  1            \\

Aware Home Rocker On  & 10            &               \\
Aware Home Rocker Off & 9             & 1             \\

Ikea Drawer Open      & 10            &               \\
Ikea Drawer Close     & 10            &               \\

Brown Drawer  Open    &  10            &               \\
Brown Drawer Close    &  10            &               \\

Orange Switch On      &  8            & 2             \\
Orange Switch Off     &  9            & 1             \\

Ornate Switch On      &  9            & 1             \\
Ornate Switch Off     &  9            & 1             \\
\hline
\end{tabular}
\end{table}

\begin{figure*}
\centering
\includegraphics[height=2.80cm]{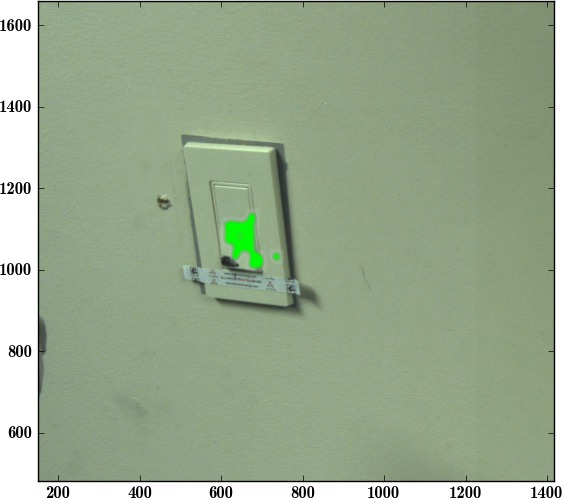}
\includegraphics[height=2.80cm]{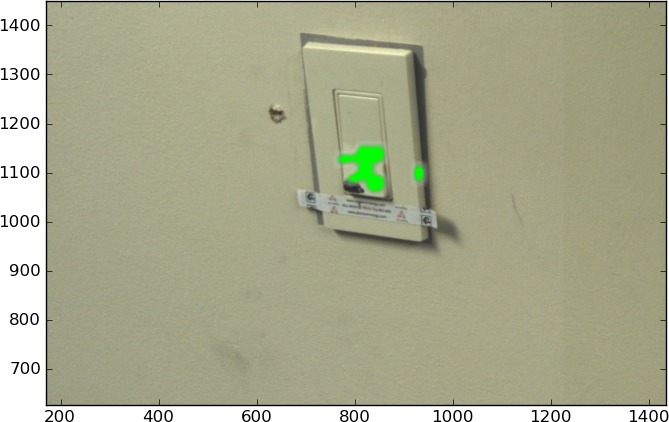}
\includegraphics[height=2.80cm]{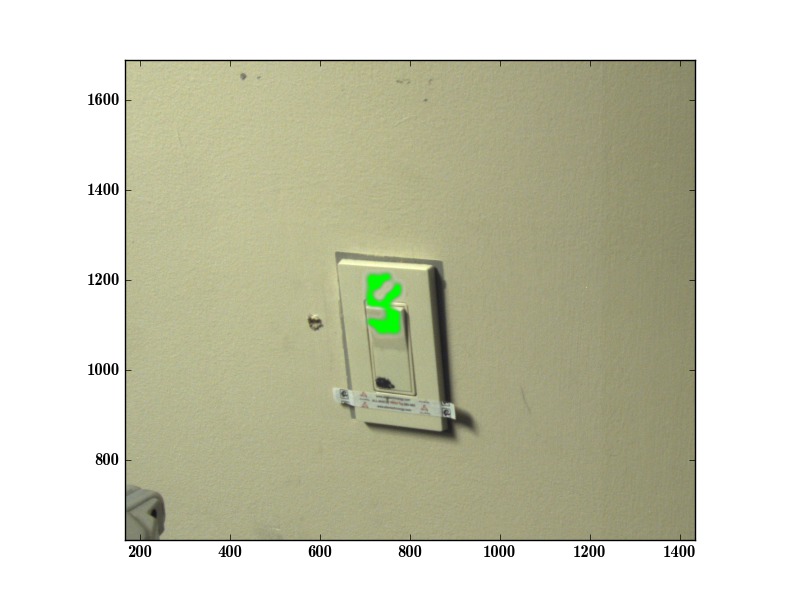}
\includegraphics[height=2.80cm]{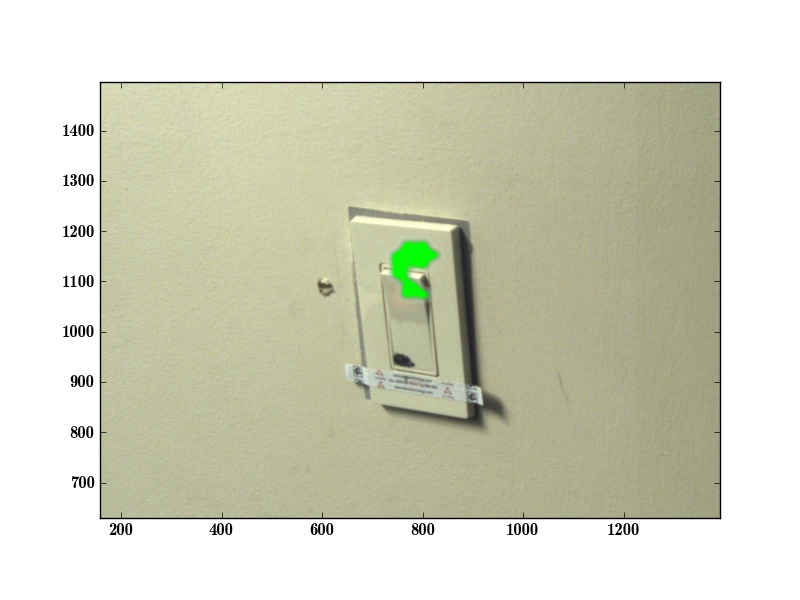}

\includegraphics[height=2.80cm]{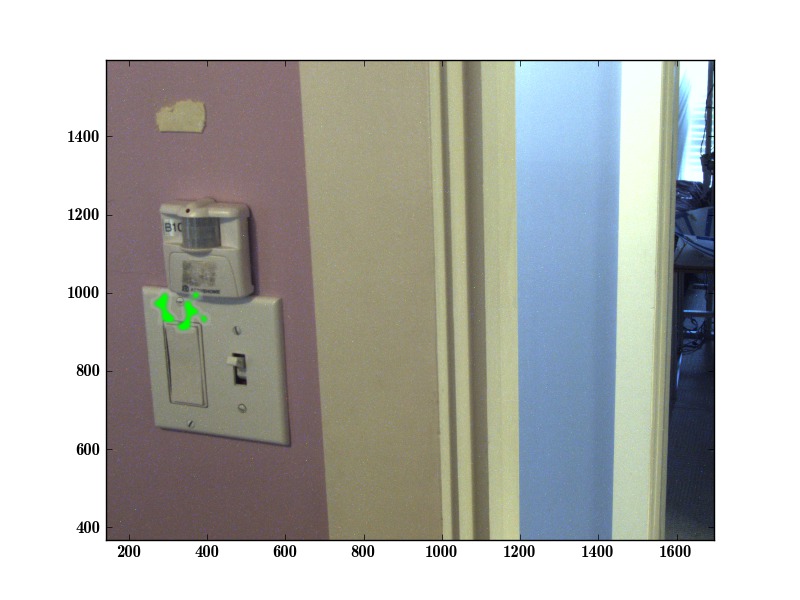}
\includegraphics[height=2.80cm]{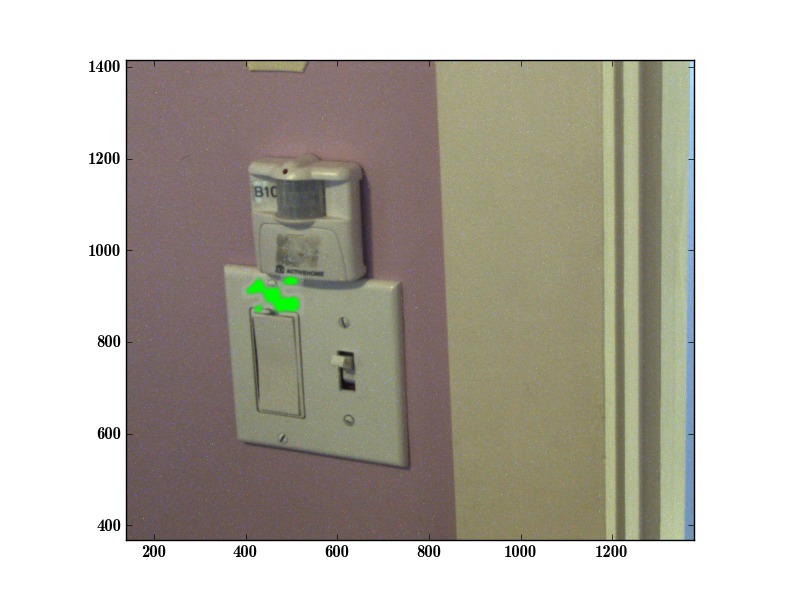}
\includegraphics[height=2.80cm]{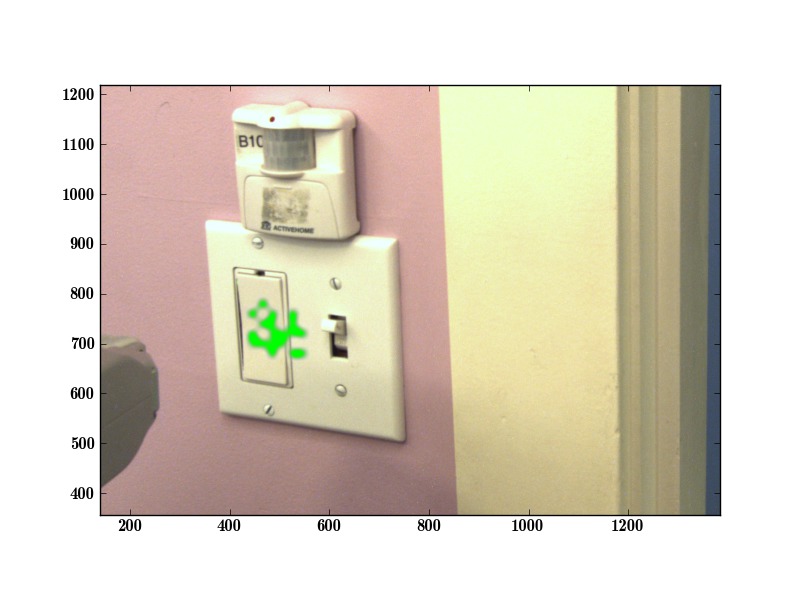}
\includegraphics[height=2.80cm]{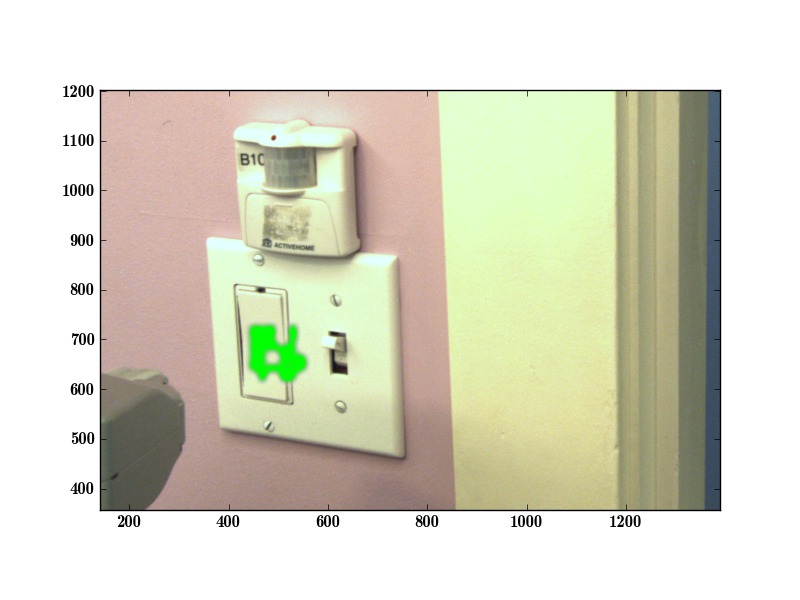}

\includegraphics[height=2.80cm]{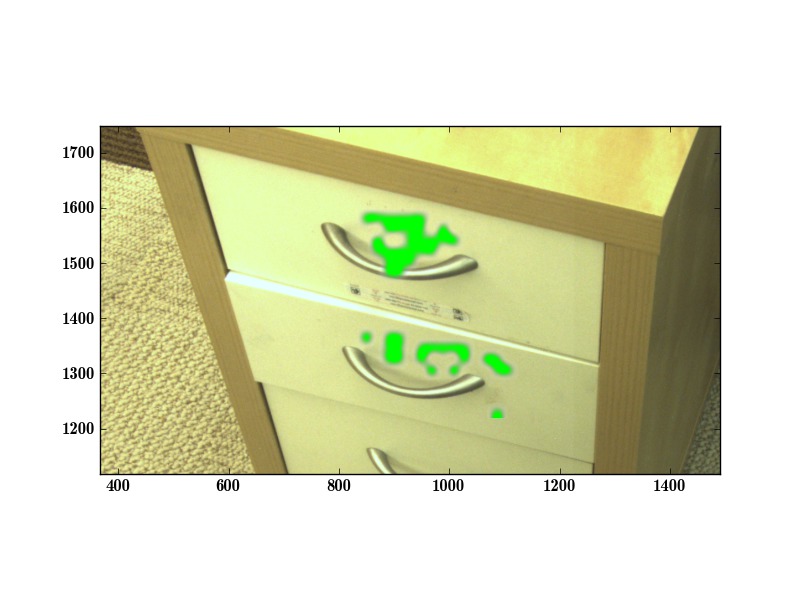}
\includegraphics[height=2.80cm]{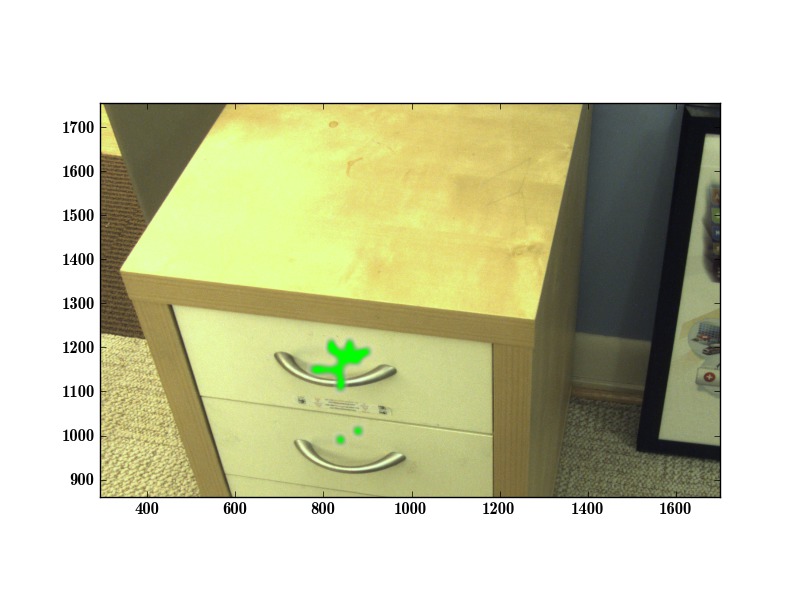}
\includegraphics[height=2.80cm]{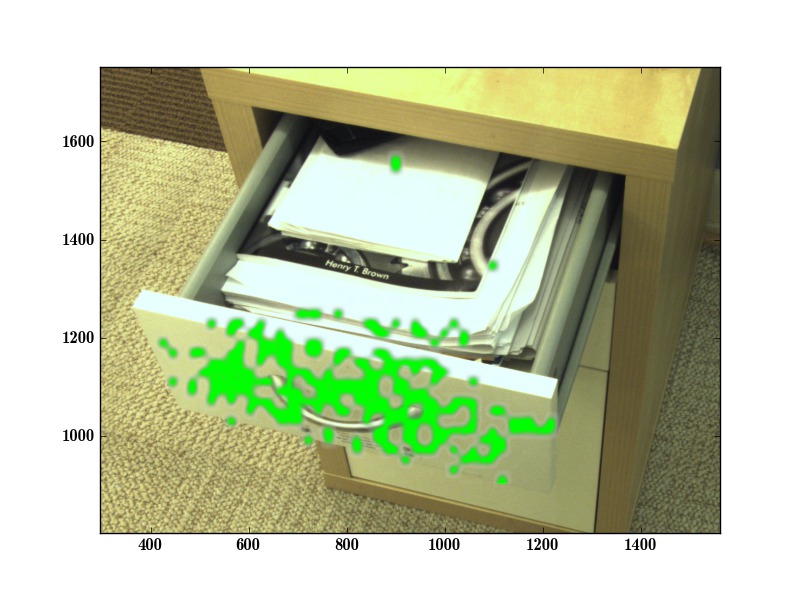}
\includegraphics[height=2.80cm]{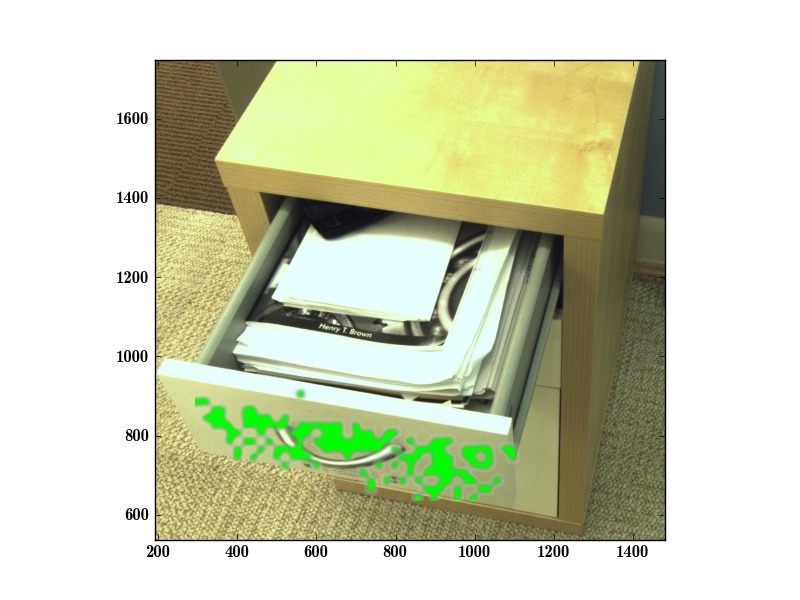}

\includegraphics[height=2.80cm]{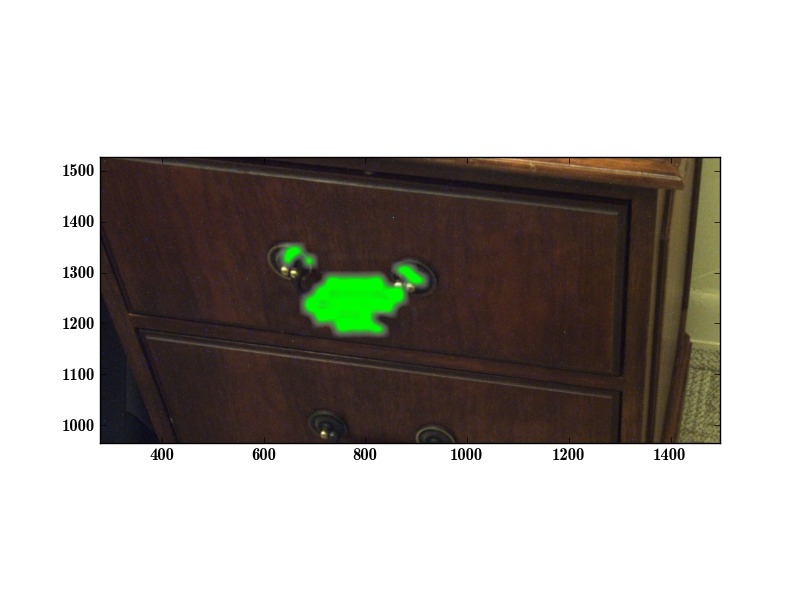}
\includegraphics[height=2.80cm]{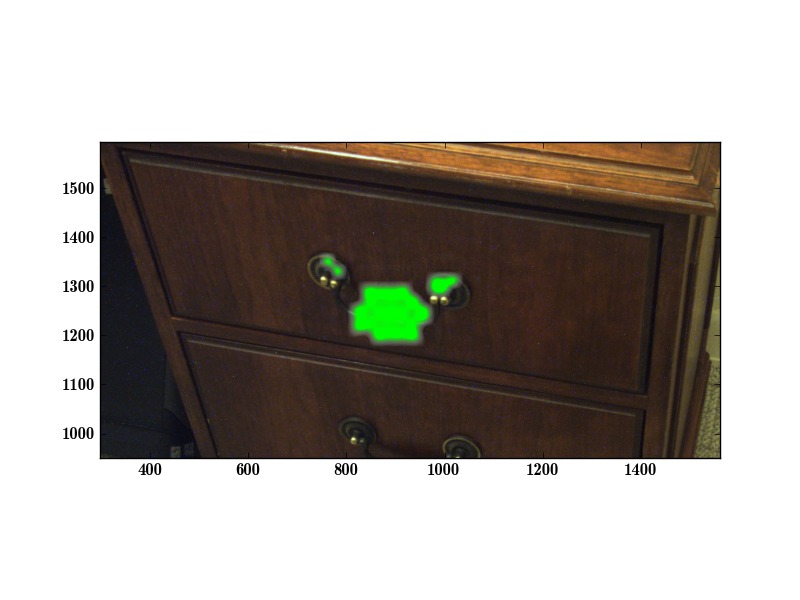}
\includegraphics[height=2.80cm]{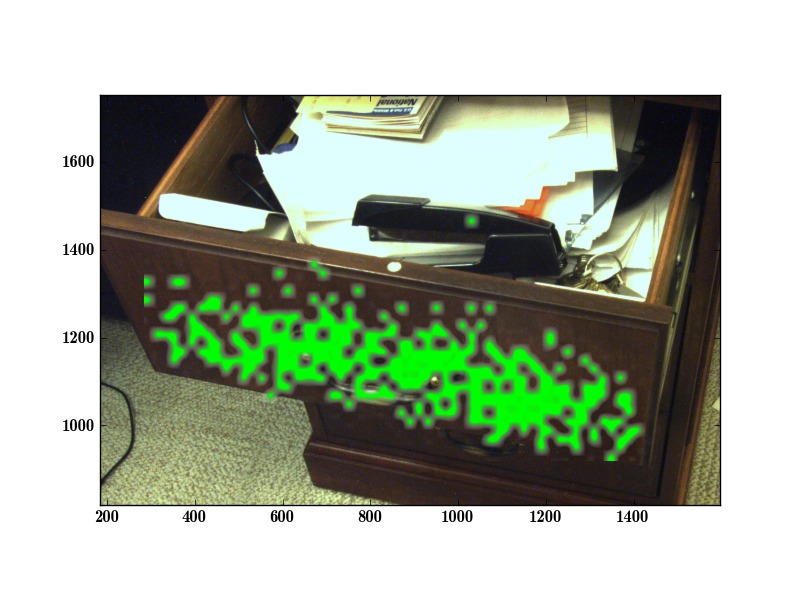}
\includegraphics[height=2.80cm]{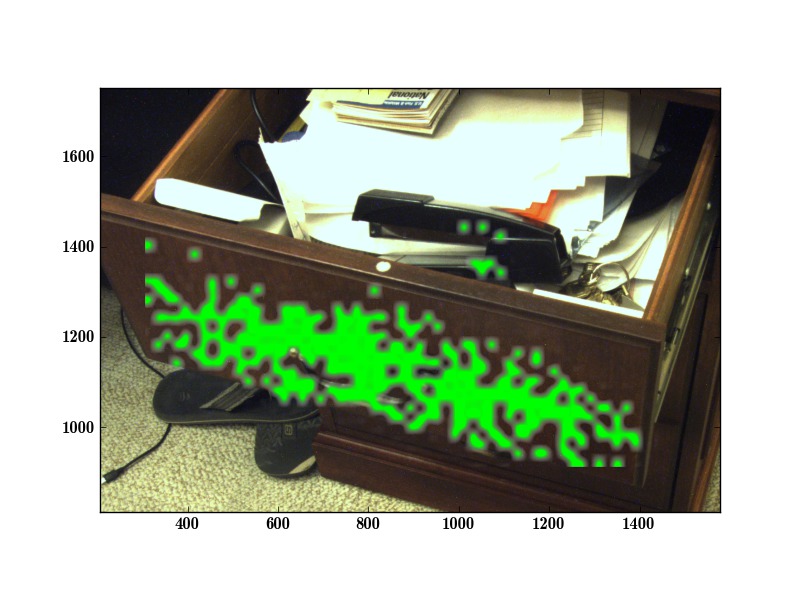}

\includegraphics[height=2.80cm]{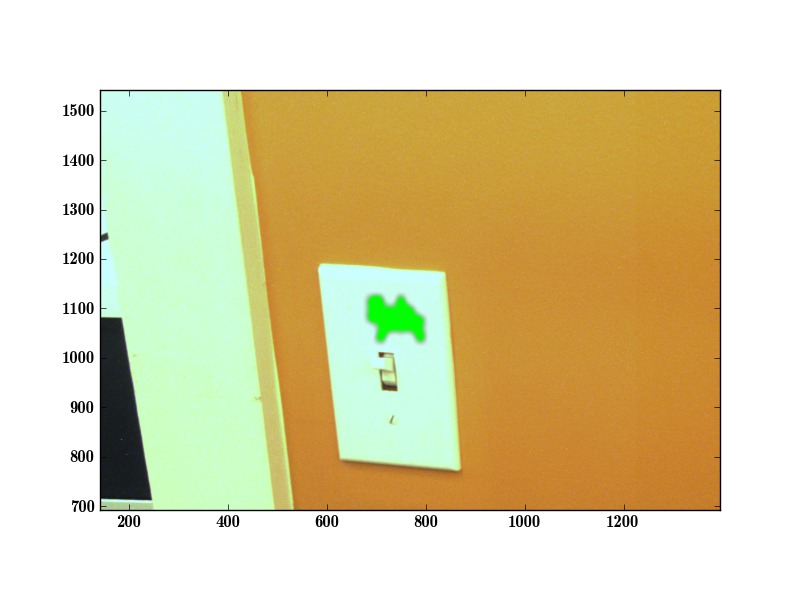}
\includegraphics[height=2.80cm]{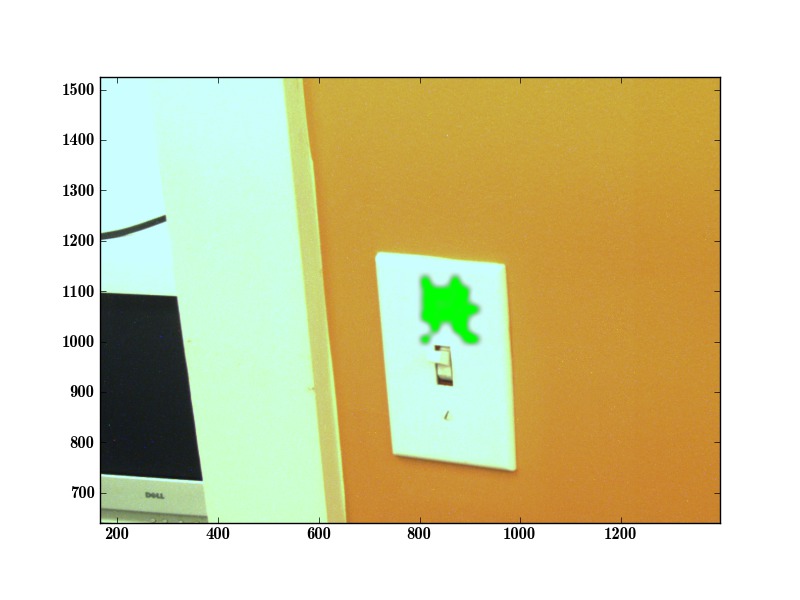}
\includegraphics[height=2.80cm]{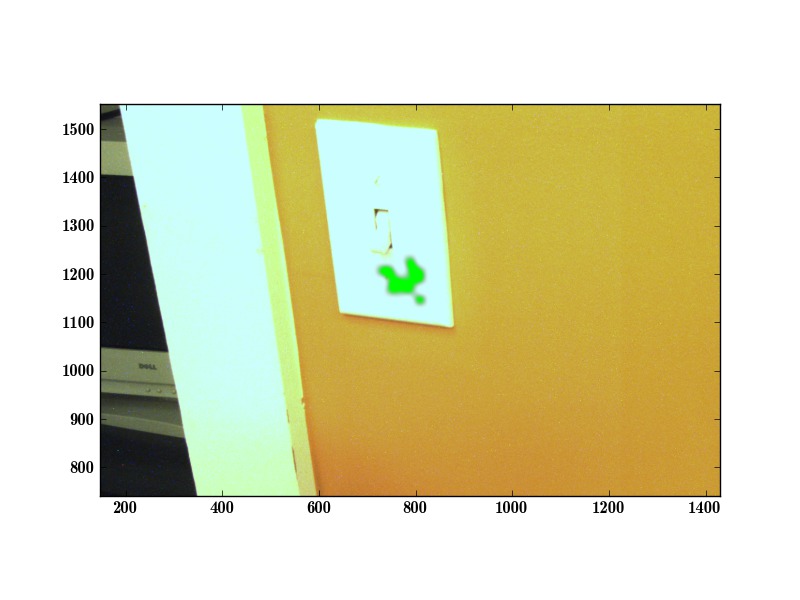}
\includegraphics[height=2.80cm]{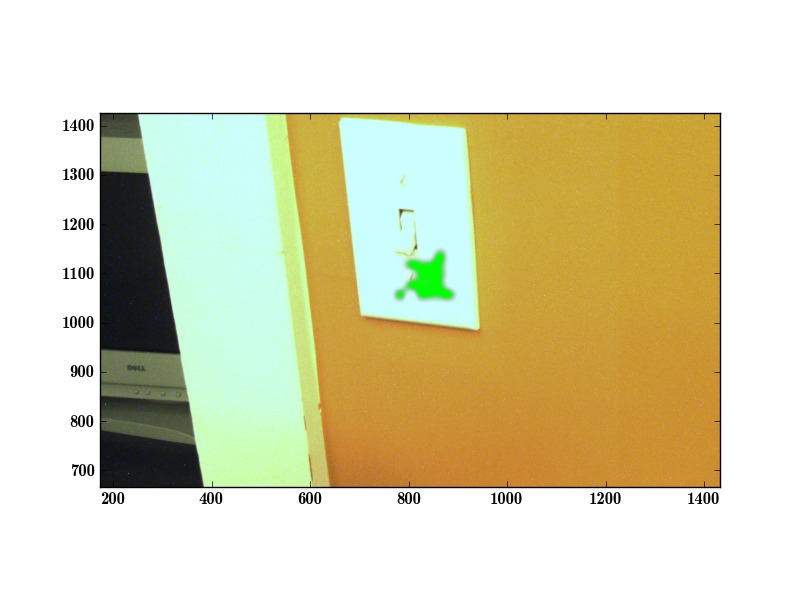}

\includegraphics[height=2.80cm]{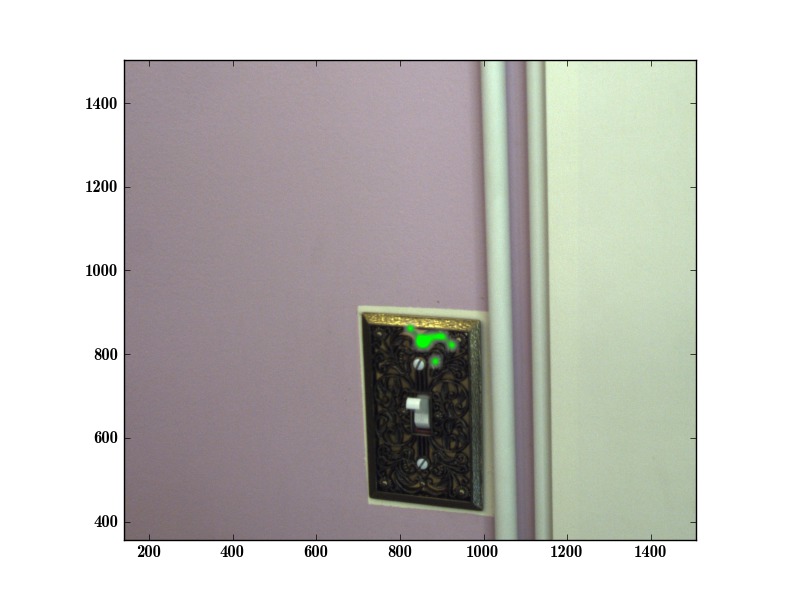}
\includegraphics[height=2.80cm]{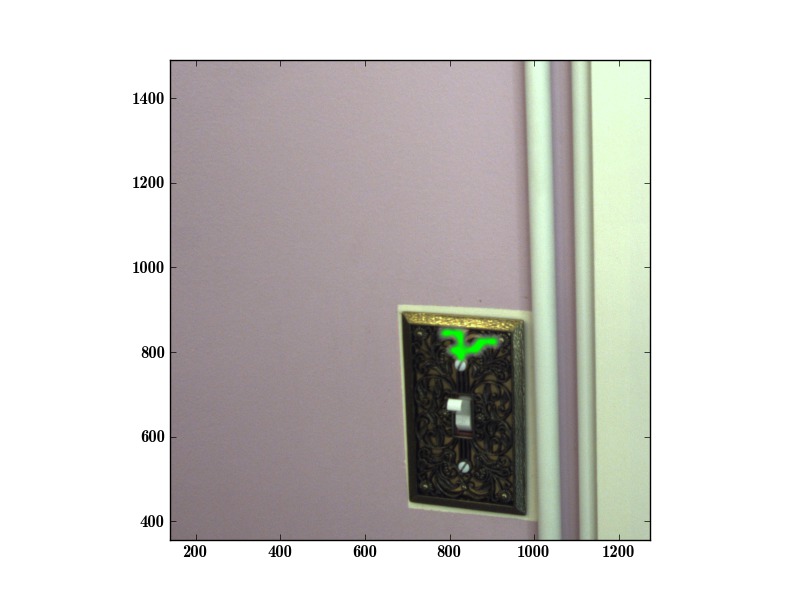}
\includegraphics[height=2.80cm]{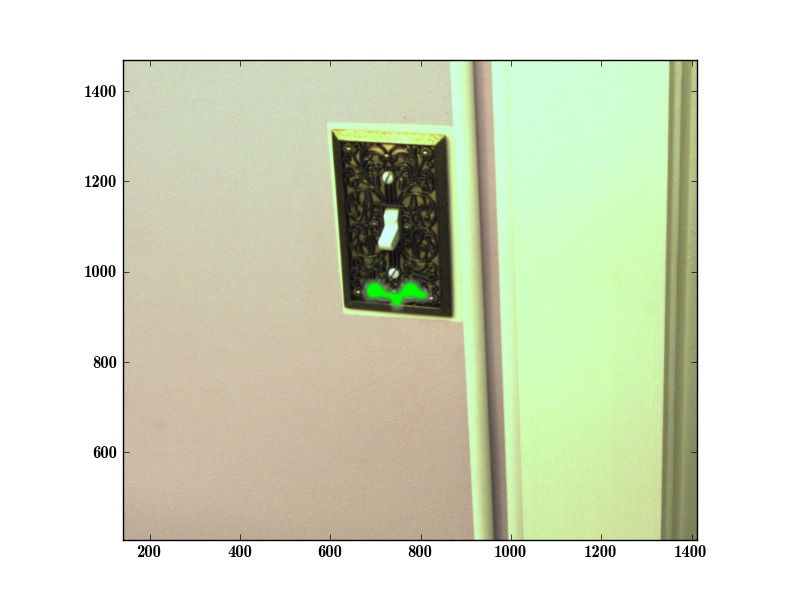}
\includegraphics[height=2.80cm]{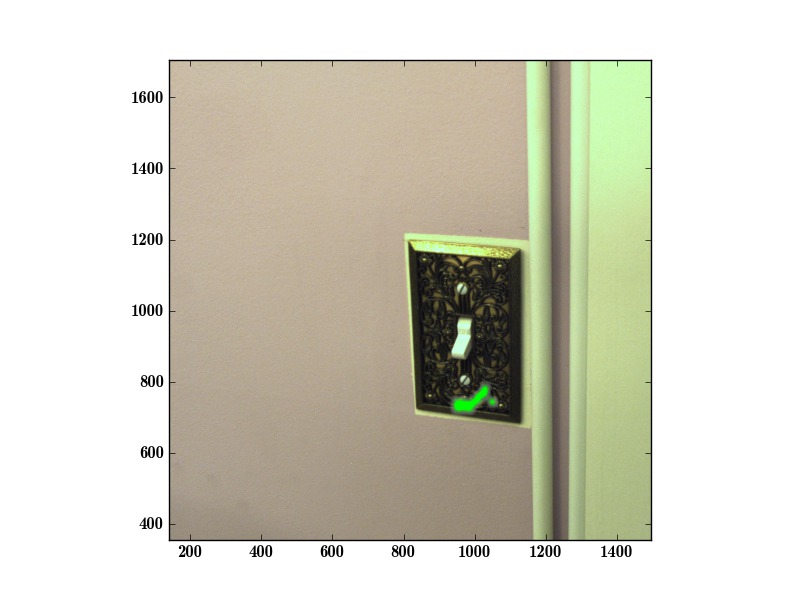}

\caption{
\label{fig_classification}
Each pair of images shows classification results of learned detectors just
after convergence then on a new test image.  Green areas mark locations
identified as leading to success of associated behaviors.
\textbf{Row 1:} Detectors for a rocker switch in our lab.
\textbf{Row 2:} Detectors for a different rocker switch in the Aware Home.
\textbf{Row 3:} Detectors for pushing and pull a wooden drawer.
\textbf{Row 4:} Detectors for another dark wooden drawer.
\textbf{Row 5:} Detectors for a regular light switch.
\textbf{Row 6:} Detectors for an ornate light switch.
}
\end{figure*}

Figure \ref{fig_classification} shows the locations that the
autonomously trained SVMs predict will be likely to lead to the
success of their associated behaviors. These predictions are solely a
function of the visual appearance of each location as represented by
its image feature vector. These visualizations of the classifier
output demonstrate that the classifiers identify locations relevant to
their associated behaviors. For example, the robot autonomously
discovers that opening a drawer requires grasping at the location of
the drawer handle, while closing a drawer can be performed across the
front surface of the drawer. The visualizations also show that
different drawer handles can have distinct task-relevant
properties. For example, the opening behavior works best when grasping
the middle of the silver handle, but can succeed by grasping the far
ends of the brass handle. 

Due to the distribution for random sampling including some points on
the lower handles for the white drawers, the SVM estimates that
success can be achieved by pulling on the top handle or the bottom
handle. The illustrates a limitation with our current approach, since
the verification function for pulling a drawer open can not tell the
difference between the top or the bottom drawer. It also shows the
influence of the distribution used to randomly sample 3D locations. At
the same time, it suggests that the visual classifiers may have some
ability to generalize to distinct objects.

For the light switches, the behaviors slide along the surface of the
switch. The robot autonomously discovered that locations that are
along the switch plate above and below the switch are likely to lead
to success.  Additionally, it does not predict success for locations
along the wall, which is appropriate since the robot's fingers get
caught on the switch plate edge if the robot tries to slide along the
wall to the switch.

In Table \ref{table_training}, we show the number of examples
collected for each classifier.  The median number of examples needed
was 77, and the maximum needed was 145 examples.  With the rocker
switch, where examples are noisy due to the middle of the switch being
an unreliable spot to push, the number of examples increased to 145
indicating a sensitivity of our approach to label noise.

Table \ref{table_execution} shows the results of using these trained
classifiers after training.  Encouragingly, over the 110 trials our
behavior execution process attained a 100\% success rate after at most
two tries.  In addition, errors that led to retries usually caused the
robot to miss an appropriate location on the device by a small
distance.

\section{Future Work}
\label{sec_limitations}

There are a number of potential extensions to this work, and
interesting issues left to consider. Although we have picked a
particular active learning framework, other frameworks might perform
better. In addition, currently we assume that each new device is
completely new to the robot, but many devices of a particular class
have visual similarities. Data from other devices might provide a
prior and reduce the training required. Similarly, the structure of
successful locations might be shared across devices, even if they are
visually distinct. For example, the front surfaces of drawers often
being pushable, the centers of drawers often being pullable, and the
centers of light switch panels often being switchable could be useful
information, even if aspects of their appearances change dramatically.

\section{Discussion and Conclusions}
\label{sec_conclusion}

In general, there are risks for a robot that learns in human
environments and an unrestrained learning system can get into
situations that are dangerous to itself, to the environment, or to
people.  We address this issue by limiting the robot to using a few
classes of behaviors in parts of the home that users have designated
as safe for robot learning. Additionally, the behaviors move the
robot's arm compliantly and use haptic sensing to decide when to stop
moving. By learning in situ, a robot's data gathering activities do
not have to stop after its training phase and can potentially continue
for as long as the robot remains in service.

Autonomous learning in human environments is a promising area of
research that gives robots methods to cope with devices that they have
not encountered before and many forms of real-world variation.  We
have presented methods that enable a mobile manipulator to
autonomously learn to visually predict where manipulation attempts
will succeed. As we discussed in the introduction, our work advances
autonomous robot learning in three ways. First, our approach uses a
robot's mobility as an integral part of autonomous learning, which
enables the robot to handle the significant task variation introduced
by its mobility. Second, our research demonstrates that by using
active learning, a robot can autonomously learn visual classifiers
solely from self-generated data in real-world scenarios with a
tractable number of examples. Third, our research introduces
complementary behaviors to address challenges associated with
autonomously learning tasks that change the state of the world.

\section*{Acknowledgments}

We thank Aaron Bobick, Jim Rehg, and Tucker Hermans for their
input. We thank Willow Garage for the use of a PR2 robot, financial
support, and other assistance. This work was funded in part by NSF
awards CBET-0932592, CNS-0958545, and IIS-1150157.

\bibliographystyle{abbrv}
\bibliography{task_relevant_feature_learning}

\end{document}